%% file: main.tex
\documentclass[Afour,sageh,times]{sagej}

\usepackage{moreverb,url}

\usepackage[colorlinks,bookmarksopen,bookmarksnumbered,citecolor=red,urlcolor=red]{hyperref}

\usepackage{graphicx}
\usepackage{float}
\usepackage{amsmath,amssymb,amsfonts}
\usepackage{amsthm}
\usepackage{mathrsfs}
\usepackage{booktabs}
\usepackage{multirow}
\usepackage{rotating}
\usepackage[labelsep=period]{caption}
\usepackage{enumitem}
\usepackage[utf8]{inputenc}
\usepackage{titlesec}
\usepackage{xcolor}
\usepackage{textcomp}
\usepackage{manyfoot}
\usepackage{algorithm}
\usepackage{algorithmicx}
\usepackage{algpseudocode}
\usepackage{listings}
\usepackage{tikz}
\usetikzlibrary{patterns, arrows.meta, shapes.geometric, arrows, fit, positioning}
\usepackage{pgfplots}
\pgfplotsset{compat=1.18}
\usepackage{microtype}
\newtheorem{definition}{Definition}

\newcommand\BibTeX{{\rmfamily B\kern-.05em \textsc{i\kern-.025em b}\kern-.08em
T\kern-.1667em\lower.7ex\hbox{E}\kern-.125emX}}

\def\volumeyear{2016}

\begin{document}

\runninghead{Siddiqui, Islam, and Noor}

\title{A Temporal Planning Approach for Intelligent Flood Response}

\author{Fazlul Hasan {Siddiqui}\affilnum{1}, Md. Monjurul {Islam}\affilnum{1,2}, and Sabah Binte {Noor}\affilnum{1}}

\affiliation{\affilnum{1}Department of Computer Science and Engineering, Dhaka University of Engineering \& Technology, Gazipur, 1707, Bangladesh
\\
\affilnum{2}Department of Computer Science and Engineering, Bangladesh Army University of Engineering \& Technology, Natore, 6431, Bangladesh
}

\corrauth{Sabah Binte {Noor}, Department of Computer Science and Engineering, Dhaka University of Engineering \& Technology, Gazipur, 1707, Bangladesh.}

\email{sabah@duet.ac.bd}

\begin{abstract}
Effective response to multiple, simultaneously flooded areas requires coordinating appropriate actions in the correct temporal order, under severe resource constraints. Automated planning provides a foundation for addressing this challenge by generating time-aware schedules, given a formal description of available resources, constraints, and goals. This work presents an intelligent flood-response framework that exploits temporal planning and models the complete operational life cycle of flood response. The framework incorporates priority-driven triage, route accessibility and travel costs, resource allocation, and supply management, while also supporting mid-execution re-planning in response to unexpected environmental changes. The framework is formulated both in the Action Notation Modeling Language (ANML) and the Planning Domain Definition Language (PDDL) 2.1, facilitating compatibility with a wider range of temporal planners.  Experimental results establish the feasibility and scalability of the proposed framework, showing that flood response scenarios can be effectively modeled and solved using temporal planning, while providing guidance on planner selection.
\end{abstract}


\maketitle

\input{introduction}
\input{background}
\input{methodology}

\input{experiment}
\input{conclusion}

\subsection*{\normalsize\sagesf\bfseries Author Contributions}
\begin{refsize}
\noindent
All authors contributed to the conceptualization of the study and the development of the research methodology. Fazlul Hasan Siddiqui secured the project funding, administered the project, provided the necessary resources, and supervised the research. Md. Monjurul Islam conducted the investigation, developed the software, performed the visualization of the results, and prepared the original draft of the manuscript. Fazlul Hasan Siddiqui and Sabah Binte Noor validated the methodology and research findings. All authors contributed to the review and editing of the manuscript and approved the final version for publication.
\end{refsize}

\section*{Statements and Declarations}

\subsection*{\normalsize\sagesf\bfseries Funding}
\begin{refsize}
\noindent
This work was supported by the University Grants Commission of Bangladesh through the Office of the Director (Research and Extension), DUET, Gazipur [grant number DUET-TRF/2025-2026/13].
\end{refsize}

\subsection*{\normalsize\sagesf\bfseries Data Availability}
\begin{refsize}
\noindent
The data supporting the findings of this study are available from the corresponding author upon reasonable request.
\end{refsize}

\subsection*{\normalsize\sagesf\bfseries Code Availability}
\begin{refsize}
\noindent
The source code used in this work is publicly available at the repository
\url{https://github.com/islammonjurul/fresched}.
\end{refsize}

\bibliographystyle{SageH}
\bibliography{bibliography}

\end{document}

%% file: introduction.tex
\section{Introduction}
An effective response to floods requires rapid decision-making and coordinated actions \citep{kapucu2011collaborative}. Traditional flood management combines structural and non-structural measures. Structural measures include embankments and improved drainage, while non-structural measures include early warning systems~\citep{meyer2012economic}. 
However, traditional approaches are increasingly considered inadequate to deal with complex flood events, especially when multiple locations are affected at the same time and response resources have to be allocated under severe time constraints~\citep{rutherford2024can}. 
Automated planning systems offer a viable solution to this problem. 

Automated planning systems use heuristic-based search to reason over a formal description of available resources and operational constraints \citep{ghallab2004automated}. 
In this work, we formulate flood response as a temporal planning problem to provide structured coordination and scheduling of response actions under resource constraints, along with dynamic replanning to adapt to environmental changes during execution of the scheduled actions.

Earlier works in disaster management focused primarily on preparedness and prevention, whereas recent studies emphasize response, recovery, and resilience. 
However, absolute protection is neither achievable nor sustainable due to inherent uncertainties and high costs~\citep{schanze2006flood}. 
Therefore, strategic flood risk management principles recommend a portfolio of diverse responses rather than relying on a single measure~\citep{sayers2013flood}. 
A comparative study of the severe 1998 and 2020 Yangtze River floods by \cite{jia2022flood} showed that the results were significantly improved in 2020 due to the advanced emergency response and risk management capabilities. This study indicates that effective response measures are critical for mitigating flood impacts and strengthening community resilience.

Several studies formalize specific flood response operations as computational problems. \cite{chang2007scenario} use stochastic programming models to optimize rescue depot locations and supply allocation under uncertainty. 
They employed a sample average approximation scheme to solve their models.  
\cite{simonovic2005computer} developed a simulation model that captures human behavior during flood emergency evacuation. 
Their model incorporates key variables such as the number of families at risk, evacuated populations, warning systems, and route inundation. 
At a broader scale, \cite{ye2021towards} propose an interdisciplinary, AI-driven framework for urban flood resilience that integrates urban planning, landscape architecture, and computer science, addressing the persistent gap between research and practice in achieving flood resilience. Whereas \cite{ambily2024framework} present a spatial planning framework centered on blue-green infrastructure for prioritizing key infrastructure dimensions.

Other researchers focus specifically on evacuation logistics during extreme events. 
\cite{yin2024strategic} examine storm flood evacuations in large coastal cities, with particular attention to elderly populations. 
Their optimization of shelter placement and routing evacuees to nearby rather than distant facilities improved overall evacuation efficiency. 
Meanwhile, \cite{wang2024pedestrian} study pedestrian evacuation during dam-break floods.  
They utilize a fuzzy VIKOR method to evaluate potential shelters and a dual-objective model to balance evacuation time, cost, and adaptability under evolving post-disaster conditions.
While these works address important aspects of flood response, none employ automated planning and scheduling to generate time-aware, coordinated response plans under resource constraints.

The most relevant work in this domain is the RAPID framework by \cite{islam2025rapid}, which introduces a disaster response domain using the Planning Domain Definition Language (PDDL) \citep{pddl} with numeric planners to generate disaster response plans. 
However, RAPID relies on classical PDDL with numeric fluents, and therefore, it cannot model action scheduling or the concurrent overlapping of operations. 
This study addresses that limitation by introducing a temporal planning approach for temporally coherent flood response.

In this work, we formalize flood response as a coordination problem to optimize critical operational decisions, including the allocation of teams, resources, and vehicles, as well as the timing and sequencing of concurrent actions. 
We present a temporal planning framework, modeled in two formal languages, the Action Notation Modeling Language (ANML) \citep{smith2008anml} and PDDL~2.1 \citep{fox2003pddl2}, covering the full operational cycle of flood response.
The proposed framework incorporates real-world constraints like priority-based triage ordering, route accessibility, vehicle capacity, and scheduling of concurrent operations. 
Additionally, the framework includes a symbolic milestone system that captures the multi-trip nature of evacuation and supply delivery without the computational overhead of continuous numeric reasoning.
Notably, the framework supports mid-execution replanning to adapt to the dynamic changes in the environment. As new data is revealed from the field, it allows us to update the instance to the current observable state and trigger replanning for an adapted schedule.
We also provide practical planner selection guidance based on scenario complexity and quality requirements.

\subsection*{Illustrative Example}
\begin{figure*}[htbp]
    \centering
    \includegraphics[clip, trim={0cm 0.5cm 0cm 0.5cm}, width=\textwidth]{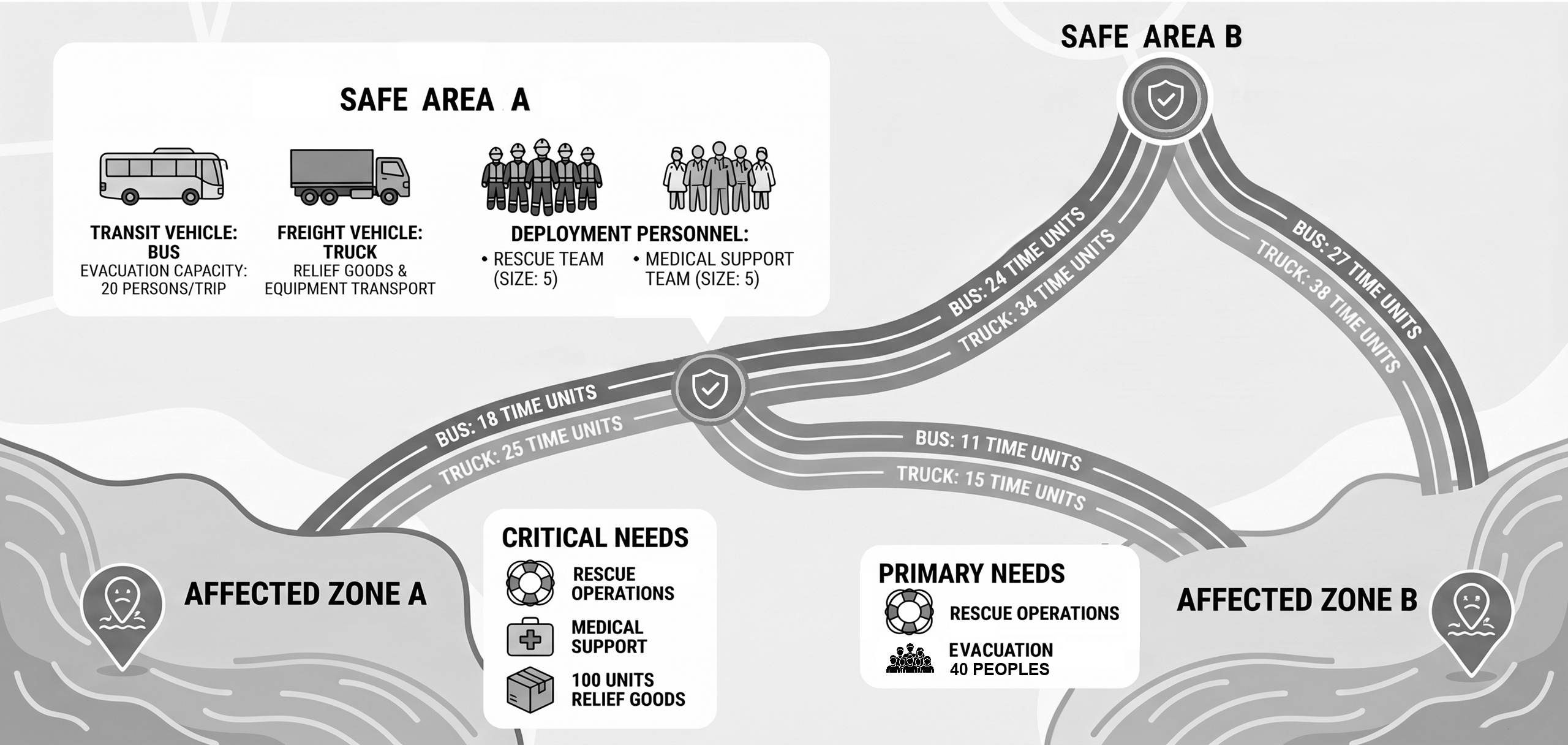}
    \caption{An example scenario with two safe areas (\texttt{safeA} and \texttt{safeB}) and two affected zones (\texttt{zoneA} and \texttt{zoneB}).
    \texttt{zoneA} requires rescue operations, medical support, and 100 units of relief goods, while \texttt{zoneB} requires rescue operations and evacuation of 40 people. Vehicle-dependent travel times range from 11 to 38 time units.
    }
    \label{fig:scenario}
\end{figure*}

A representative flood response scenario used to demonstrate the proposed domain is shown in Figure~\ref{fig:scenario}. 
The scenario consists of two safe areas, \texttt{safeA} and \texttt{safeB}, and two affected areas, \texttt{zoneA} and \texttt{zoneB}.
The location \texttt{safeA} is the main deployment base that has a transit vehicle (\texttt{bus}) with an evacuation capacity of 20 persons per trip, a freight vehicle (truck) used for resource transportation, a rescue team of 5 members, a medical support team of 5 members, and 100 units of relief goods (food). 

Affected area \texttt{zoneA} requires rescue operations, medical support, and 100 units of food, whereas \texttt{zoneB} requires rescue operations and evacuation of 40 people. 
Travel time between locations depends on the vehicle. For instance, a bus takes 18 time units to move from \texttt{safeA} to \texttt{zoneA}, while a truck takes 25, consistent with real-world cases where vehicle type determines travel time.

\begin{figure*}[htb]
    \centering
    \begin{tikzpicture}[
        x=0.044cm, y=0.5cm,
        font=\sffamily,
        transport/.style={fill=white},
        response/.style={fill=black!15},
        logistics/.style={fill=white,
            pattern=north east lines,
            pattern color=black!40}
    ]
    
    \begin{scope}
    \foreach \y in {1, 3, ..., 25} {
        \fill[gray!7] (-110, -\y+0.5) rectangle (285, -\y-0.5);
    }
    \end{scope}
    
    \foreach \x in {0, 25, 50, 75, 100, 125, 150, 175, 200, 225, 250, 275} {
        \draw[gray!30, thin] (\x, -0.5) -- (\x, -25.5);
        \node[text=gray!70!black, font=\fontsize{7}{8}\selectfont] at (\x, 0) {\x};
    }
    \foreach \x in {5,10,15,20,30,35,40,45,55,60,65,70,80,85,90,95,105,110,115,120,130,135,140,145,155,160,165,170,180,185,190,195,205,210,215,220,230,235,240,245,255,260,265,270} {
        \draw[gray!15, thin] (\x, -0.5) -- (\x, -25.5);
    }
    \draw[gray!10, thin] (-110, -0.5) -- (-110, -25.5);
    \draw[gray!10, thin] (285, -0.5) -- (285, -25.5);
    \draw[thick, gray!70!black, ->] (0, -0.5) -- (285, -0.5) node[below left, font=\fontsize{7}{8}\selectfont] {Time~~~~~};
    
    \newcommand{\ganttbar}[5]{
        \node[anchor=east, font=\fontsize{7}{8}\selectfont, text=black!90] at (-2, -#3) {#2};
        \draw[#1, draw=black!60, thin, rounded corners=1pt] (#4, -#3+0.35) rectangle (#4+#5, -#3-0.35);
        \node[font=\tiny, text=black] at (#4+#5/2, -#3) {#5};
    }
    
    \ganttbar{logistics}{LoadResource(food,safeA,truck,k100,k50)}{1}{0}{50}
    \ganttbar{transport}{BoardTeam(medics,safeA,bus)}{2}{0}{5}
    \ganttbar{transport}{BoardTeam(rescuers,safeA,bus)}{3}{0}{5}
    \ganttbar{transport}{MoveVehicle(bus,safeA,zoneA)}{4}{5}{18}
    \ganttbar{transport}{DisembarkTeam(medics,bus,zoneA)}{5}{23}{5}
    \ganttbar{transport}{DisembarkTeam(rescuers,bus,zoneA)}{6}{23}{5}
    \ganttbar{response}{RescueAffectedPeople(zoneA,rescuers)}{7}{28}{15}
    \ganttbar{response}{ProvideMedicalSupport(zoneA,medics)}{8}{28}{15}
    \ganttbar{transport}{BoardTeam(rescuers,zoneA,bus)}{9}{43}{5}
    \ganttbar{transport}{MoveVehicle(bus,zoneA,zoneB)}{10}{48}{29}
    \ganttbar{transport}{MoveVehicle(truck,safeA,zoneA)}{11}{50}{25}
    \ganttbar{logistics}{DeliverResource(food,zoneA,truck,k100,k50)}{12}{75}{50}
    \ganttbar{logistics}{Evacuate(zoneB,bus,p40,p20)}{13}{77}{20}
    \ganttbar{transport}{DisembarkTeam(rescuers,bus,zoneB)}{14}{77}{5}
    \ganttbar{response}{RescueAffectedPeople(zoneB,rescuers)}{15}{82}{15}
    \ganttbar{transport}{MoveVehicle(bus,zoneB,safeA)}{16}{97}{11}
    \ganttbar{logistics}{DisembarkEvacuatedPeople(bus,safeA)}{17}{108}{10}
    \ganttbar{transport}{MoveVehicle(bus,safeA,zoneB)}{18}{118}{11}
    \ganttbar{transport}{MoveVehicle(truck,zoneA,safeA)}{19}{125}{25}
    \ganttbar{logistics}{Evacuate(zoneB,bus,p20,p0)}{20}{129}{20}
    \ganttbar{transport}{MoveVehicle(bus,zoneB,safeA)}{21}{149}{11}
    \ganttbar{logistics}{LoadResource(food,safeA,truck,k50,k0)}{22}{150}{50}
    \ganttbar{logistics}{DisembarkEvacuatedPeople(bus,safeA)}{23}{160}{10}
    \ganttbar{transport}{MoveVehicle(truck,safeA,zoneA)}{24}{200}{25}
    \ganttbar{logistics}{DeliverResource(food,zoneA,truck,k50,k0)}{25}{225}{50}
    
    \begin{scope}[shift={(0, 0.75)}]
        \fill[transport, draw=black!60, thin, rounded corners=1pt] (0,0) rectangle (8, 0.6); 
        \node[right, font=\fontsize{7}{8}\selectfont] at (10, 0.3) {Transport};
        
        \fill[response, draw=black!60, thin, rounded corners=1pt] (75,0) rectangle (83, 0.6); 
        \node[right, font=\fontsize{7}{8}\selectfont] at (85, 0.3) {Response};
        
        \fill[logistics, draw=black!60, thin, rounded corners=1pt] (150,0) rectangle (158, 0.6); 
        \node[right, font=\fontsize{7}{8}\selectfont] at (160, 0.3) {Logistics};
    \end{scope}
    
    \end{tikzpicture}
    \caption{A temporal plan for the example flood response scenario in Figure~\ref{fig:scenario}. Each bar represents a durative action, where the horizontal placement indicates the start time and the width with the inscribed number indicates the execution duration in time units. Actions aligned vertically execute concurrently.}
    \label{fig:plan}
\end{figure*}
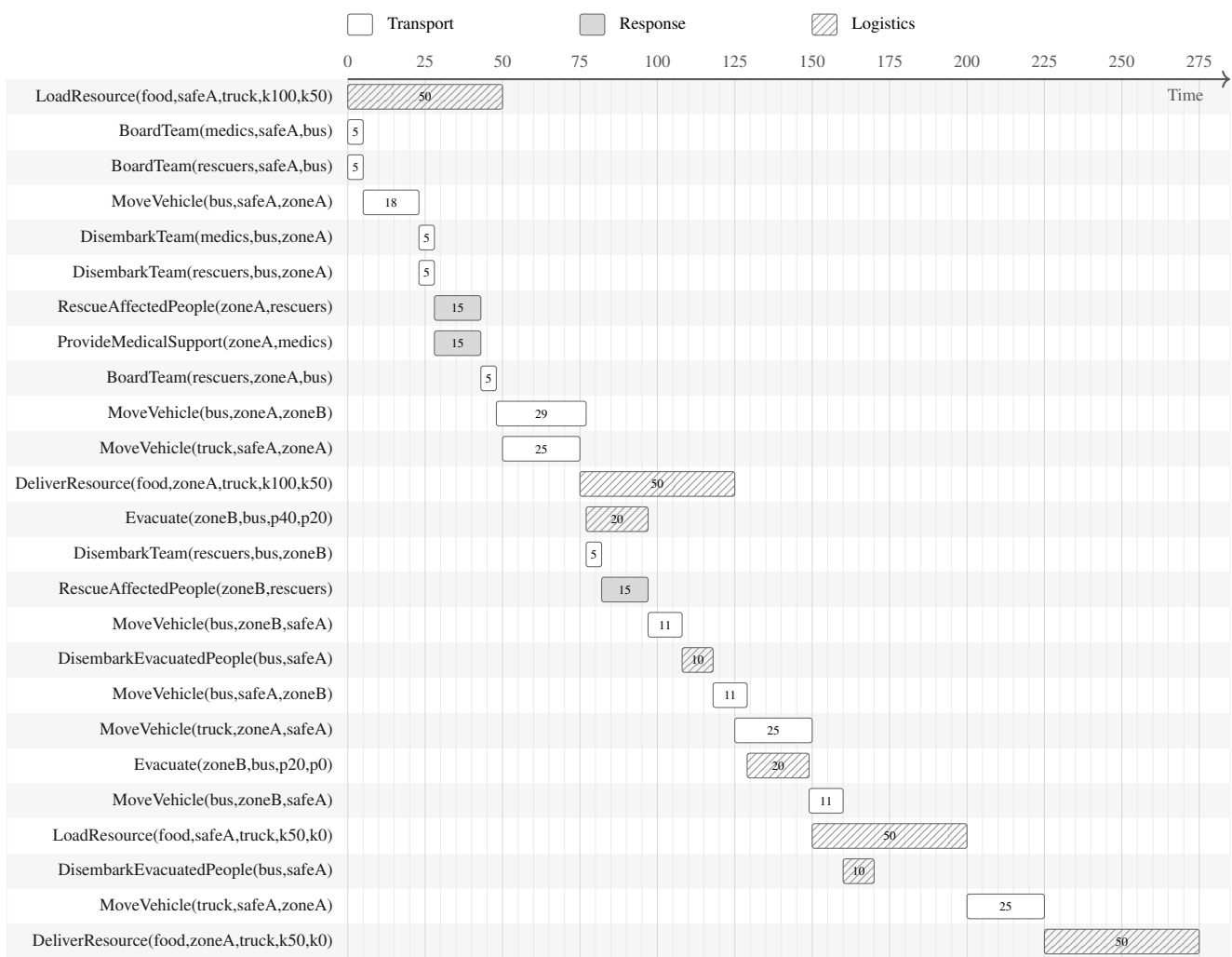

To illustrate how temporal planning operates on such a scenario, Figure~\ref{fig:plan} shows a valid action schedule, produced by an automated temporal planner.
The plan consists of 25 actions with a makespan of 275 time units. 
This plan demonstrates how temporal planning coordinates overlapping operations. Specifically, at time~0, three actions execute concurrently: loading relief goods into the truck (\texttt{LoadResource}), while the medical support and rescue teams simultaneously board the bus (\texttt{BoardTeam}). 
After that, the bus departs at time~5 and arrives at \texttt{zoneA} at time~23, and both teams disembark from the bus and immediately begin rescue operations (\texttt{RescueAffectedPeople}) and medical support (\texttt{ProvideMedicalSupport}) in parallel at time~28. 

In addition, the truck departs at time~50 and arrives at \texttt{zoneA} at time~75, and relief goods are then delivered. 
Meanwhile, the rescue team is transported and is redeployed to \texttt{zoneB} via bus, where evacuation and rescue operations are carried out across two evacuation sorties (\texttt{Evacuate} at times~77 and~129). Each evacuation sortie is followed by a return trip to \texttt{safeA} to disembark evacuees before redeployment. 
Applying the whole plan efficiently resolves all demands for evacuation, rescue, medical support, and resource delivery across the impacted zones, and several actions concurrently take place, similar to a real-world scheduling scenario.

%% file: background.tex
\section{Background}
In this section, we introduce basic notions of automated planning and modeling languages used in this work. We first define the primary automated planning paradigms, including classical, numeric, and temporal planning, and then present an overview of the two modeling languages, PDDL and ANML, used to model the proposed domain.

\subsection{Automated Planning}
Automated planning generates sequences of actions (i.e., plans) which transform an initial state into a desired goal state~\citep{ghallab2004automated}. Depending on the complexity of the environment of the problem, planning models need to reason about logical conditions, quantitative resources, and temporal constraints. These requirements give rise to three progressively expressive paradigms: classical planning, numeric planning, and temporal planning.

Classical planning (Definition~\ref{def:classic}) provides the fundamental structural syntax for automated planning. It operates within a deterministic, fully observable, and static environment, meaning the agent has complete knowledge of the system state and the deterministic outcomes of its actions. 

\begin{definition}
    \label{def:classic}
    A \textbf{classical planning task} is represented as a tuple $\Pi = \langle F, A, \gamma, I, G \rangle$, where:
    \begin{itemize}
        \item $F$ is a finite set of propositional facts,
        \item $A$ is a finite set of actions,
        \item $\gamma: A \mapsto \mathbb{Q}_0^+$ is a function assigning a non-negative rational cost to each action,
        \item $I \subseteq F$ represents the initial state, and
        \item $G \subseteq F$ specifies the goal conditions.
    \end{itemize}
    Each action $a \in A$ is a tuple $a=\langle pre(a), add(a), del(a) \rangle$, where $pre(a), add(a), del(a) \subseteq F$ denote the action's precondition, add list, and delete list, respectively. The add and delete lists collectively constitute the action effects.
\end{definition}

A solution to a classical planning task is a sequence of actions that transforms the initial state into a state satisfying the goal conditions. An action is applicable only when its preconditions hold in the current state, after which its effects modify the state by adding and removing facts. Definition ~\ref{def:plan} formally defines a \textbf{plan}.

\begin{definition}
    \label{def:plan}
    A \textbf{plan} $\pi$ for a classical planning task $\Pi$ is an ordered sequence of actions $\langle a_1, a_2, \dots, a_n \rangle$. The plan $\pi$ is valid if and only if $pre({a_1}) \subseteq I$ and the sequential application of actions $a_1, a_2, \dots, a_n$ successfully yields a state satisfying the goal conditions $G$.
\end{definition}
Classical planning cannot represent quantitative resources such as fuel, supply quantities, or vehicle capacities because states are purely propositional. Many real-world problems require reasoning about quantities that change during execution. Numeric planning directly extends classical planning by introducing numeric state variables and arithmetic expressions into the planning model \citep{fox2003pddl2}. 
\begin{definition}
\label{def:numeric}
A \textbf{numeric planning task} is defined as a tuple $\Pi_n = \langle F, X, A, \gamma, I, G \rangle$, where $F$, $A$, and $\gamma$ retain their definitions from the classical planning paradigm (Definition~\ref{def:classic}), and   $X$ is a finite set of numeric variables bounded over the rational numbers $\mathbb{Q}$, $I \subseteq F\cup X$ is the initial state, and $G \subseteq F\cup X$ specifies the goal conditions.
\end{definition}

Actions of a numeric planning task may modify both propositional facts and numeric variables through arithmetic updates.
Despite this added expressiveness, numeric planning still assumes that actions occur instantaneously. Real-world operations involve activities with duration, deadlines, and concurrency requirements that numeric planning tasks fail to encode. Temporal planning addresses these limitations by explicitly incorporating time into the planning process  \citep{fox2003pddl2}.

\begin{definition}
\label{def:temporal}
A \textbf{temporal planning task} is represented as a tuple
$\Pi_t = \langle F, X, A^d, \gamma, I, G \rangle$,
where $F$, $X$, $\gamma$, $I$, and $G$ are defined as in a
numeric planning task (Definition~\ref{def:numeric}), and
$A^d$ is a finite set of \emph{durative actions}.
Each action $a \in A^d$ is a tuple as given in \eqref{eq:durativeaction}.
\begin{equation}
\label{eq:durativeaction}
  a = \langle\,
    \mathit{pre}^s(a),\;
    \mathit{pre}^o(a),\;
    \mathit{pre}^e(a),\;
    \mathit{eff}^s(a),\;
    \mathit{eff}^e(a),\;
    \mathit{dur}(a)
  \,\rangle
\end{equation}
where $\mathit{pre}^s(a)$, $\mathit{pre}^o(a)$, $\mathit{pre}^e(a)$ $\subseteq F \cup X$ are the \emph{at-start}, \emph{over-all}, and \emph{at-end} preconditions respectively, $\mathit{eff}^s(a)$, $\mathit{eff}^e(a)$ $\subseteq F\cup X$ are the \emph{at-start} and \emph{at-end} effects, and $\mathit{dur}(a) \in \mathbb{R}^+$ is the action duration.

\end{definition}
Temporal planning extends numeric planning by modeling actions as durative activities whose conditions and effects may occur at different points during execution. At-start preconditions must hold at the moment execution begins, over-all preconditions must remain true throughout the open interval of the action's duration, and at-end preconditions must hold upon completion.
Effects are applied at their respective time points, modifying the state by adding and removing facts. This enables planners to reason about overlapping actions, synchronization constraints, and time-dependent resource usage. Additionally, Timed Initial Literals (TILs) allow facts to change independently at predefined time points.

The solution to a temporal planning task is a timed schedule of actions, i.e., a temporal plan (Definition~\ref{def:temporalplan}).
\begin{definition}
    \label{def:temporalplan}
    A \textbf{temporal plan} $\pi_t$ for a temporal planning task $\Pi_t$ is a timed schedule of durative actions represented as a tuple  $\langle(t_{1}: a_1), (t_{2}: a_2), ..., (t_n: a_n)\rangle$, where each $t_i \in \mathbb{R}^+_0$ is a time point of the start of some durative action $a_i \in A_d$. 
    The schedule $\pi_t$ is valid iff all action preconditions are satisfied at their required time points, and the goal conditions are achieved upon completion of the schedule.
\end{definition}
Plan quality in temporal planning is most commonly measured by \emph{makespan} (Definition~\ref{def:makespan}), the total time required to execute a schedule.
\begin{definition}
\label{def:makespan}
The \textbf{makespan} of a temporal plan $\pi_t = \langle(t_1{:}a_1),(t_2{:}a_2),\dots,(t_n{:}a_n)\rangle$ is the elapsed time between the start of its earliest action and the completion of its latest action, as specified in \eqref{eq:makespan}. A smaller
makespan reflects a more temporally compact schedule with greater
concurrency among actions.
\begin{equation}
\label{eq:makespan}
\mathrm{makespan}(\pi_t) \;=\;
\max_{1\le i\le n}\bigl(t_i + \mathrm{dur}(a_i)\bigr)
\;-\; \min_{1\le i\le n} (t_i)
\end{equation}
\end{definition}

Flood response operations require synchronized activities with durations involving constrained time and resources, so temporal planning is the natural fit for this work.

\subsection{Planning Modeling Languages}

Planning modeling languages specify planning domains and problem instances in terms that automated planners can process: defining actions, state variables, constraints, temporal properties, initial states, and goal conditions. Early languages worked with propositional representations and instantaneous actions; modern ones also handle quantitative reasoning, temporal constraints, durative actions, and concurrency. This work uses two temporal planning languages, PDDL and ANML, and the concepts from each that are relevant here are the following.

\subsubsection{The Planning Domain Definition Language}
The Planning Domain Definition Language (PDDL) is a standardized formalism for modeling automated planning domains and problems \citep{pddl}. It separates reusable domains containing typed predicates, functions, and parameterised action schemas from instance-specific information such as the objects, the initial state, and the goal.
PDDL~2.1 extends the classical propositional language with \emph{numeric fluents} and \emph{durative actions} \citep{fox2003pddl2}, enabling quantitative and temporal reasoning. 

A durative action in PDDL~2.1 specifies a duration together with temporally annotated conditions and effects. 
Three qualifiers locate these in time: \texttt{at start} conditions and effects apply at the instant execution begins, \texttt{over all} conditions must hold throughout the open interval of the action, and \texttt{at end} conditions and effects apply on completion.
The Temporal Fast Downward (TFD) planner used in our evaluation consumes PDDL~2.1.

\subsubsection{The Action Notation Modeling Language} The Action Notation Modeling Language (ANML) is a high-level formalism for modeling planning domains and problems \citep{smith2008anml}. 
Where PDDL is primarily propositional, ANML adopts a variable/value representation in which every variable is an implicit function of time, so that state transitions over an action's execution are expressed directly on a timeline. 
Types follow a single-inheritance hierarchy and may carry member variables and constants.

Conditions and effects are not separated by keywords but distinguished by their operators: relational operators (e.g.\ \texttt{==}) express conditions, while assignment and transition operators (e.g.\ \texttt{:=} and \texttt{:->}) express effects. Each condition and effect is scoped by a temporal qualifier relative to the action's \texttt{start} and \texttt{end} landmarks, and \texttt{all} denotes the whole execution window. 
The Flexible Action and Planning Environment (FAPE) used in our evaluation consumes ANML.

%% file: methodology.tex
\section{Proposed Temporal Planning Framework}

The proposed framework for flood-response operations models disaster-response activities as a temporal planning task involving heterogeneous response teams, transportation assets, operational priorities, and resource-constrained logistics under time-dependent execution. Its objective is to generate temporally consistent schedules that coordinate rescue operations, medical assistance, civilian evacuation, and humanitarian supply distribution across geographically dispersed flood-affected areas.
The framework comprises three main components: a \emph{domain definition} that specifies the state representation, constraints, and durative actions of the flood-response environment; an \emph{instance formulation} that defines a specific disaster scenario through its initial and goal states; and \emph{plan generation}, in which an automated temporal planner generates a valid execution schedule. The remainder of this section describes each component.

\begin{figure*}[htbp]
\centering
\begin{tikzpicture}[
  loc/.style={draw, rounded corners, minimum width=2.4cm,
              minimum height=1.15cm, align=center, font=\scriptsize},
  safe/.style={loc, fill=black!4},
  zone/.style={loc, fill=black!9},
  rt/.style={-{Stealth[length=5pt]}, thin},
  lbl/.style={font=\scriptsize}
]
  \node[safe] (s) at (0,0) {\textbf{Safe $S$}\\ TransitVehicle $T$\\ FreightVehicle $F$\\ RescueTeam $R$\\MedicalSupportTeam $M$\\ \texttt{resource\_available(food)}=\texttt{k300}};
  \node[zone] (a) at (6,2) {\textbf{Zone $A$} -- priority~1\\
  \texttt{evac\_level}=\texttt{p60}\\
  \texttt{needs\_rescue}\\ \texttt{needs\_medical\_support}\\ \texttt{resource\_needed(food)}=\texttt{k200}};
  \node[zone] (b) at (6,-2) {\textbf{Zone $B$} -- priority~2\\ \texttt{evac\_level}=\texttt{p20}};
  \draw[rt] (s.40) to[bend left=30] node[lbl, above, sloped]{\texttt{T}:$20$, \texttt{F}:$25$} (a.170);
  \draw[rt] (s.320) to[bend right=30] node[lbl, below, sloped]{\texttt{T}:$30$, \texttt{F}:$40$} (b.190);
  \draw[-{Stealth[length=5pt]}, dashed] (a) to[bend left=20] node[lbl, right]{\texttt{prior} (serve first)} (b);
\end{tikzpicture}
\caption{A compact flood-response instance used as a running example. A single safe location $S$ stages a transit vehicle $T$, a freight vehicle $F$, a rescue team $R$, and a medical support team $M$, together with a stock of relief goods (\texttt{food}=\texttt{k300}). Zone $A$ outranks Zone $B$ via \texttt{prior}. Edge labels give vehicle-specific travel times.}
\label{fig:example}
\end{figure*}
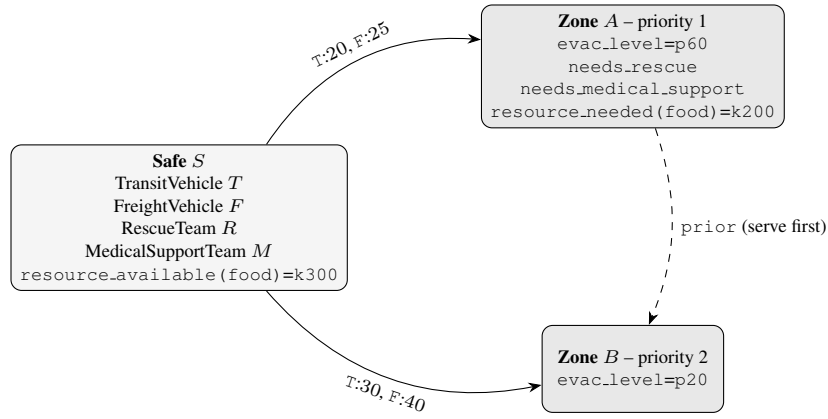

We use a compact illustrative scenario, presented in Figure~\ref{fig:example}, as a running example throughout this section.
A single safe location $S$ serves as a staging hub containing a transit vehicle $T$, a freight vehicle $F$, a rescue team $R$, a medical team $M$, and a stock of relief goods (e.g., \texttt{food}=\texttt{k300}). From this safe hub, operations are carried out over two affected zones: Zone $A$ (priority 1) and Zone $B$ (priority 2). Each zone has associated evacuation and relief requirements. A strict triage constraint is imposed through the \texttt{prior} relation: each response operation in Zone $A$ must be completed before the same operation in Zone $B$ can begin. Vehicle-specific travel times between the safe hub and affected zones are encoded as edge labels. This compact instance captures all core structural elements of the domain: shared vehicles, competing zones, temporal routing constraints, and resource-dependent service execution.

\subsection{Task Formulation}
\label{sec:formulation}

We present the proposed model using a propositional temporal-planning formulation (Definition~\ref{def:flood}), following the general formulation introduced in Definition~\ref{def:temporal}. We first characterize the state space, then the action model, and finally the validity conditions a solution plan must satisfy. The flood-response planning task is stated as follows.

\begin{definition}
\label{def:flood}
A \textbf{flood-response planning task} is a temporal planning problem $\Pi_f = \langle F_f,\, X_f,\, A_f^d,\, \gamma_f,\, I_f,\, G_f \rangle$
defined over a flood-response domain, where:
\begin{itemize}[leftmargin=*]

\item $F_f$ is the finite set of grounded propositional facts over the predicate set $\mathcal{P}$, specified in \eqref{eq:predicate}.

\begin{equation}
\label{eq:predicate}
\mathcal{P} =
\left\{
\begin{aligned}
&\texttt{prior},
 \texttt{vehicle\_at},
 \texttt{team\_at},\\
 &\texttt{team\_in\_vehicle},
 \texttt{evac\_level},\\
 &\texttt{needs\_rescue},
 \texttt{resource\_needed},\\
 &\texttt{needs\_medical\_support},
 \texttt{evacuating},\\
 &\texttt{resource\_available},
 \texttt{cargo\_loaded},\\
 &\texttt{cargo\_full},
 \texttt{route\_accessible},\\
 &\texttt{valid\_evac\_step},
 \texttt{valid\_supply\_step}
\end{aligned}
\right\}
\end{equation}

\item $X_f$ is the finite set of ground numeric fluents (all
\emph{rigid}, i.e.\ never modified by an action effect) over the set $\mathcal{X}$ = \{\texttt{travel\_time}, \texttt{team\_size}, \texttt{person\_count\_diff}, \texttt{package\_count\_diff}\}.

\item $A_f^d$ is the finite set of ground durative actions over the set of nine action schemas $\mathcal{A}$, specified in \eqref{eq:actions}, instantiated over the typed objects of a given problem instance.

\begin{equation}
\label{eq:actions}
\mathcal{A} =
\left\{
\begin{aligned}
&\textsc{BoardTeam},
 \textsc{DisembarkTeam},\\
& \textsc{MoveVehicle},
 \textsc{Evacuate},\\
&\textsc{DisembarkEvacuatedPeople},\\
&\textsc{RescueAffectedPeople},\\
&\textsc{LoadResource},
 \textsc{DeliverResource},\\
 &\textsc{ProvideMedicalSupport}
\end{aligned}
\right\}
\end{equation}

\item $\gamma_f$ is the action cost function, where $\gamma_f = 1$ for all $a \in A_f^d$.

\item $I_f \subseteq F_f\cup X_f$ is the initial state, grounded over $\mathcal{P}\cup\mathcal{X}$ to describe the initial configuration of vehicles, teams, routes, operational requirements, and milestone levels for a given flood scenario.

\item $G_f \subseteq F_f$ is the goal condition grounded over $\mathcal{P}$, as formalized in \eqref{eq:goal}.
\begin{equation}
\label{eq:goal}
G_f =
\left\{
\begin{aligned}
& \neg\texttt{needs\_rescue}(l),
\neg\texttt{evacuating}(v),\\
&\neg\texttt{needs\_medical\_support}(l),\\
 &\texttt{evac\_level}(l,\texttt{p0}),\\
&\texttt{resource\_needed}(l,r,\texttt{k0})
\end{aligned}   
\right\}
\end{equation}
for all $l \in \mathcal{L}_A$, $r \in \mathcal{R}$, and $v \in \mathcal{V}_T$,
where $\mathcal{L}_A$ denotes the set of affected locations, $\mathcal{R}$ is the set of resource types, $\mathcal{V}_T$ is the set of transit vehicles, and $\texttt{p0}$ and $\texttt{k0}$ are the base milestone values denoting full evacuation and full resource satisfaction.
\end{itemize}
\end{definition}

\subsection{Modeling Decisions}
\label{sec:design}

The following modeling decisions are intended to balance representational expressiveness with planning tractability and motivate the predicates and action schemas introduced in the following subsections.

\paragraph{\textbf{Symbolic milestones instead of numeric fluents.}}
Evacuation and supply progress are represented using ordered symbolic milestones rather than continuous numeric fluents. This design avoids large numeric search spaces while preserving multi-trip operational behavior and capacity-aware progress tracking.

\paragraph{\textbf{Priority-constrained execution.}}
Area-specific priorities are encoded directly within the preconditions of rescue, medical support, evacuation, and resource-delivery actions. Consequently, every generated plan automatically respects operational priorities without requiring additional validation or post-processing.

\paragraph{\textbf{Temporal resource locking.}}
Vehicles and teams are treated as temporally exclusive resources. During transportation, they are temporarily unavailable for other actions, thereby preventing conflicting concurrent assignments.

\paragraph{\textbf{Native concurrency.}}
The temporal-planning formalism naturally permits concurrent execution of independent rescue, medical, transportation, and logistics operations when resource and ordering constraints are satisfied.

\subsection{World Representation}
\label{sec:world}

This subsection defines the representational structure of the planning domain, including the type hierarchy, the state predicates, and the symbolic milestone system used to model operational progress.

\subsubsection{Type System}
\label{sec:types}
The proposed domain uses a typed object hierarchy to organize operational entities involved in planning. The domain is defined over four root types: \texttt{Location}, \texttt{Vehicle}, \texttt{Team}, and \texttt{Resource}. These root types are further specialized, as presented in \eqref{eq:typing}, to capture the operational structure of flood-response activities as follows:
\begin{equation}
\label{eq:typing}
\begin{aligned}
\texttt{Location}  &\supseteq
  \{\texttt{AffectedLocation},
      \texttt{SafeLocation}\} \\
\texttt{Vehicle}   &\supseteq
  \{\texttt{TransitVehicle},
      \texttt{FreightVehicle}\} \\
\texttt{Team}      &\supseteq
  \{\texttt{RescueTeam},
      \texttt{MedicalSupportTeam}\}
\end{aligned}
\end{equation}
\begin{itemize}[leftmargin=*]
    \item \textbf{Safe areas and affected zones:} The \texttt{Location} type represents all areas within the scenario and is specialized into two distinct subtypes: \texttt{AffectedLocation} and \texttt{SafeLocation}. 
    \item \textbf{Transportation Assets:} The \texttt{Vehicle} type models transportation assets and is specialized into two subtypes: \texttt{TransitVehicle} for personnel and evacuee transport and \texttt{FreightVehicle} for relief cargo delivery.
    \item \textbf{Response Teams:} The \texttt{Team} type represents deployable response units and is partitioned into two subtypes: \texttt{RescueTeam} and \texttt{MedicalSupportTeam}.
    \item \textbf{Consumable Resources:} \texttt{Resource} denotes consumable humanitarian supplies such as food, water, and medical kits. Its dynamic quantities are represented through symbolic milestones and state predicates across locations.
\end{itemize}
A distinguished constant \texttt{NULL} represents temporary unassignment of entities during transitional operations such as movement, boarding, loading, and unloading.

\subsubsection{State Representation}
\label{sec:predicates}

The predicates collectively define the
operational world state of the flood-response environment. They are organized into spatial configuration, vehicle status, location requirements, operational progress, and rigid constants.
\begin{itemize}[leftmargin=*]
    \item \textbf{Spatial configuration:} The spatial information of vehicles and teams is captured by \texttt{vehicle\_at(v,l)}, \texttt{team\_at(t,l)}, and \texttt{team\_in\_vehicle(t,v)}, denoting vehicle locations, team deployment, and team boarding status, respectively.
    \item \textbf{Vehicle status:} Vehicle operations are captured by \texttt{evacuating(v)}, \texttt{route\_accessible(v,l1,l2)}, \texttt{cargo\_full(v)}, and \texttt{cargo\_loaded(v,r)}, indicating evacuation activity, cargo status, and route feasibility.
    
    \item \textbf{Location requirements and priority:} Operational requirements are expressed using \texttt{needs\_rescue(l)} and \texttt{needs\_medical\_support(l)}. A priority relation \texttt{prior(l,pl)} enforces that location \texttt{pl} must be served before \texttt{l}, restricting concurrent servicing of lower-priority areas.

    \item \textbf{Operational progress and milestones:} Evacuation and supply delivery progress are modeled using ordered symbolic milestones rather than numeric fluents. For instance, two ordered domains can be defined: \texttt{PersonCount} with $\texttt{p60} \succ \texttt{p40} \succ \texttt{p20} \succ \texttt{p0}$, and \texttt{PackageCount} with $\texttt{k200} \succ \texttt{k100} \succ \texttt{k0}$, where $\texttt{p0}$ and $\texttt{k0}$ denote goal satisfaction. Three predicates operate over \texttt{evac\_level(l,p)} (remaining evacuees), \texttt{resource\_needed(l,r,k)} (remaining demand), and \texttt{resource\_available(l,r,k)} (available stock at safe locations). Each relevant action advances the corresponding milestone toward its terminal.

    \item \textbf{Constants:} A set of numeric fluents governs action durations and remains fixed for each instance. Numeric fluent \texttt{team\_size(t)} defines team's boarding and disembarkation durations, \texttt{travel\_time(v,l1,l2)} defines vehicle movement durations, while \texttt{person\_count\_diff(p\_1,p\_2)} and \texttt{package\_count\_diff(k\_1,k\_2)} scale evacuation and delivery durations according to corresponding workload. Vehicle's capacity constraints are enforced via \texttt{valid\_evac\_step(v,p\_1,p\_2)} and \texttt{valid\_supply\_step(v,k\_1,k\_2)}, which restrict allowed milestone transitions.
\end{itemize}

\subsection{Action Schemas}
\label{sec:actions}
The proposed domain is defined over nine durative actions. A set of preconditions and effects specifies each action, annotated using temporal qualifiers that indicate when they apply: \emph{at start} (execution onset), \emph{over all} (maintained throughout execution), and \emph{at end} (completion).  The complete action schemas are presented in Tables~\ref{tab:transport}--\ref{tab:logistics}.

\renewcommand{\arraystretch}{1.4}
\setlength{\tabcolsep}{6pt}
\newcommand{\tf}[1]{\texttt{#1}}
\newcommand{\nt}[1]{$\neg$\texttt{#1}}
\newcommand{\Ps}{\textit{Start}}
\newcommand{\Po}{\textit{Over-all}}
\newcommand{\Es}{\textit{Start}}
\newcommand{\Ee}{\textit{End}}

\subsubsection{Transport Actions}
The transport actions, \textsc{BoardTeam}, \textsc{DisembarkTeam}, and \textsc{MoveVehicle} (Table~\ref{tab:transport}), govern the movement of teams and vehicles across the locations. \textsc{BoardTeam} and \textsc{DisembarkTeam} are symmetric operations that transfer a team into and out of a vehicle while maintaining the vehicle's location as an invariant throughout execution. Both actions scale in duration with \texttt{team\_size}.

\textsc{MoveVehicle} models physical relocation between locations and requires route accessibility between origin and destination. The vehicle remains in transit for the entire duration, and travel time is parameterized by the specific vehicle and route. Multi-hop travel is achieved by consecutive move actions through intermediate locations.

\begin{table*}[!htbp]
\centering
\caption{Transport action schemas. Parameter shorthands: $t$ denotes a team; $l$, $from$, and $to$ denote locations; $v_T$ denotes a transit vehicle; and $T$ denotes a vehicle.}
\label{tab:transport}
\footnotesize
\begin{tabular}{
  @{} p{4cm}@{}
      p{1.3cm}@{}
      p{4.5cm}@{}
      p{0.8cm}@{}
      p{3.5cm}@{}
}
\toprule
\textbf{Action}
  & \multicolumn{2}{c}{\textbf{Preconditions}}
  & \multicolumn{2}{c}{\textbf{Effects}}\\
\cmidrule(r){2-3}\cmidrule{4-5}
\tf{[Duration]}
  & \textit{ann.}
  & \textit{predicate(s)}
  & \textit{ann.}
  & \textit{predicate(s)}\\
\midrule

\textsc{BoardTeam} $(t,l,v_T)$
  & \Ps    & \tf{team\_at}$(t,l)$
  & \Es    & \nt{team\_at}$(t,l)$ \\
\tf{[team\_size}$(t)$\tf{]}
  & \Po    & \tf{vehicle\_at}$(v_T,l)$,\newline \nt{evacuating}$(v_T)$
  & \Ee    & \tf{team\_in\_vehicle}$(t,v_T)$\\
\midrule

\textsc{DisembarkTeam} $(t,v_T,l)$
  & \Ps    & \tf{team\_in\_vehicle}$(t,v_T)$
  & \Es    & \nt{team\_in\_vehicle}$(t,v_T)$ \\
\tf{[team\_size}$(t)$\tf{]} 
  & \Po    & \tf{vehicle\_at}$(v_T,l)$
  & \Ee    & \tf{team\_at}$(t,l)$\\
\midrule

\textsc{MoveVehicle} $(v,from,to)$
  & \Ps    & \tf{vehicle\_at}$(v,from)$
  & \Es    & \nt{vehicle\_at}$(v,from)$\\
\tf{[travel\_time}$(v,from,to)$\tf{]}
  & \Po    & \tf{route\_accessible}$(v,from,to)$
  & \Ee    & \tf{vehicle\_at}$(v,to)$\\
\bottomrule
\end{tabular}
\end{table*}
\subsubsection{Response Actions}
Response actions are \textsc{RescueAffectedPeople} and \textsc{ProvideMedicalSupport} (Table~\ref{tab:response}). These actions are executed on-site by dedicated teams and require that the corresponding team remains present throughout execution as an \emph{over all} invariant.
Both actions are governed by a shared triage constraint expressed through the \texttt{prior} predicate, which prevents service at a location unless all higher-priority locations have already been fully serviced for the same task type. For instance,  Zone $A$ must be cleared before Zone $B$ in Figure~\ref{fig:example}. Each action has a fixed duration of constant time units and, upon completion, removes the corresponding demand predicate at the location. Because rescue and medical teams are distinct resources, these actions may be executed concurrently at the same location when available.

\begin{table*}[!htbp]
\centering
\caption{Response action schemas. Parameter shorthands: 
$l_A$ denotes an affected location; $pl$ denotes a higher-priority affected location; and $C$ denotes a constant value.}
\label{tab:response}
\footnotesize
\begin{tabular}{
  @{} p{3.6cm}@{}
      p{1.3cm}@{}
      p{4.5cm}@{}
      p{0.8cm}@{}
      p{4.2cm}@{}
}
\toprule
\textbf{Action}
  & \multicolumn{2}{c}{\textbf{Preconditions}}
  & \multicolumn{2}{c}{\textbf{Effects}}\\
\cmidrule(r){2-3}\cmidrule{4-5}
  & \textit{ann.}
  & \textit{predicate(s)}
  & \textit{ann.}
  & \textit{predicate(s)}\\
\midrule

\multirow{2}{=}{\textsc{RescueAffectedPeople} $(l_A,rescuers,pl)$}
  & \Ps  & \tf{prior}$(l_A,pl)$,\newline \nt{needs\_rescue}$(pl)$, \newline \tf{needs\_rescue}$(l_A)$
  & --  & --\\
  & \Po  & \tf{team\_at}$(rescuers,l_A)$
  & \Ee  & \nt{needs\_rescue}$(l_A)$ \\
\midrule

\multirow{2}{=}{\textsc{ProvideMedicalSupport} $(l_A,medics,pl)$}
  & \Ps  & \tf{prior}$(l_A,pl)$,\newline \nt{needs\_medical\_support}$(pl)$,\newline \tf{needs\_medical\_support}$(l_A)$
  & --  & --\\
  & \Po  & \tf{team\_at}$(medics,l_A)$
  & \Ee  & \nt{needs\_medical\_support}$(l_A)$ \\

\bottomrule
\end{tabular}
\end{table*}

\begin{table*}[!htbp]
\centering
\caption{Logistics action schemas. Parameter shorthands: 
$l_A$ denotes an affected location; $v_T$ denotes a transit vehicle; $pf$ denotes the initial evacuation milestone; $pt$ denotes the final evacuation milestone; $pl$ denotes a higher-priority affected location than $l_A$; $l_S$ denotes a safe location; $r$ denotes a resource; $v_F$ denotes a freight vehicle; $kf$ denotes the initial resource milestone; $kt$ denotes the final resource milestone; and $C$ denotes a constant value.}
\label{tab:logistics}
\small
\begin{tabular}{
  @{} p{5cm}@{}
      p{1.5cm}@{}
      p{5cm}@{}
      p{1cm}@{}
      p{4.8cm}@{}
}
\toprule
\textbf{Action}
  & \multicolumn{2}{c}{\textbf{Preconditions}}
  & \multicolumn{2}{c}{\textbf{Effects}}\\
\cmidrule(r){2-3}\cmidrule{4-5}
\tf{[Duration]}
  & \textit{ann.}
  & \textit{predicate(s)}
  & \textit{ann.}
  & \textit{predicate(s)}\\
\midrule

\textsc{Evacuate} $(l_A,v_T,pf,pt,pl)$
  & \Ps
  & \tf{prior}$(l_A,pl)$,\newline
    \tf{evac\_level}$(pl,\texttt{p0})$,\newline
    \tf{evac\_level}$(l_A,pf)$,\newline
    \nt{evacuating}$(v_T)$
  & $\Es^{\dagger}$
  & \nt{evac\_level}$(l_A,pf)$,\newline
    \tf{evac\_level}$(l_A,pt)$,\newline
    \tf{evacuating}$(v_T)$\\
\tf{[person\_count\_diff}$(pf,pt)$\tf{]}
  & \Po
  & \tf{vehicle\_at}$(v_T,l_A)$,\newline
    \tf{valid\_evac\_step}$(v_T,pf,pt)$
  & --  & --\\
\midrule

\textsc{DisembarkEvacuated-People} $(v_T,l_S)$
  & \Ps  & \tf{evacuating}$(v_T)$
  & --  & --\\
\tf{[C]}
  & \Po  & \tf{vehicle\_at}$(v_T,l_S)$
  & \Ee  & \nt{evacuating}$(v_T)$\\
\midrule

\textsc{LoadResource} $(r,l,v_F,kf,kt)$
  & \Ps
  & \tf{resource\_available}$(l,r,kf)$,\newline
    \nt{cargo\_loaded}$(v_F,r)$,\newline
    \nt{cargo\_full}$(v_F)$
  & \Es
  & \nt{resource\_available}$(l,r,kf)$,\newline
    \tf{resource\_available}$(l,r,kt)$\\
\tf{[package\_count\_diff}$(kf,kt)$\tf{]}
  & \Po  
  & \tf{vehicle\_at}$(v_F,l)$,\newline
    \tf{valid\_supply\_step}$(v_F,kf,kt)$
  & \Ee  
  & \tf{cargo\_loaded}$(v_F,r)$,\newline
    \tf{cargo\_full}$(v_F)$\\
\midrule

\textsc{DeliverResource} $(r,l_A,v_F,kf,kt,pl)$
  & \Ps 
  & \tf{prior}$(l_A,pl)$,\newline
    \tf{evac\_level}$(pl,\texttt{p0})$,\newline
    \tf{resource\_needed}$(pl,r,\texttt{k0})$,\newline
    \tf{resource\_needed}$(l_A,r,kf)$,\newline
    \tf{cargo\_loaded}$(v_F,r)$
  & \Es  & \nt{resource\_needed}$(l_A,r,kf)$\\
\tf{[package\_count\_diff}$(kf,kt)$\tf{]}
  & \Po
  & \tf{vehicle\_at}$(v_F,l_A)$,\newline
    \tf{valid\_supply\_step}$(v_F,kf,kt)$
  & \Ee
  & \tf{resource\_needed}$(l_A,r,kt)$,\newline
    \nt{cargo\_loaded}$(v_F,r)$,\newline
    \nt{cargo\_full}$(v_F)$\\
\bottomrule
\multicolumn{5}{l}{\footnotesize{$^\dagger$ Effect applied at action start, enabling concurrent scheduling.}}
\end{tabular}
\end{table*}

\subsubsection{Logistics Actions}
Logistics actions (Table~\ref{tab:logistics}) implement evacuation and supply operations across the same spatial network under shared priority constraints. \textsc{Evacuate} moves civilians in discrete capacity-bounded steps governed by \texttt{valid\_evac\_step}, updating evacuation milestones and marking vehicles as actively evacuating. To enable concurrency, key milestone updates occur at action start.
\textsc{DisembarkEvacuatedPeople} drops off evacuees at safe locations and releases vehicles for reuse. 

Supply operations are handled by \textsc{LoadResource} and \textsc{DeliverResource}, which shuttle goods between depots and affected zones while updating resource demand milestones. Both evacuation and supply actions use discretized progress variables rather than arithmetic quantities, ensuring compatibility with symbolic planning representations and enabling repeated execution until completion.

\subsection{Dependency Among Action Schemas}
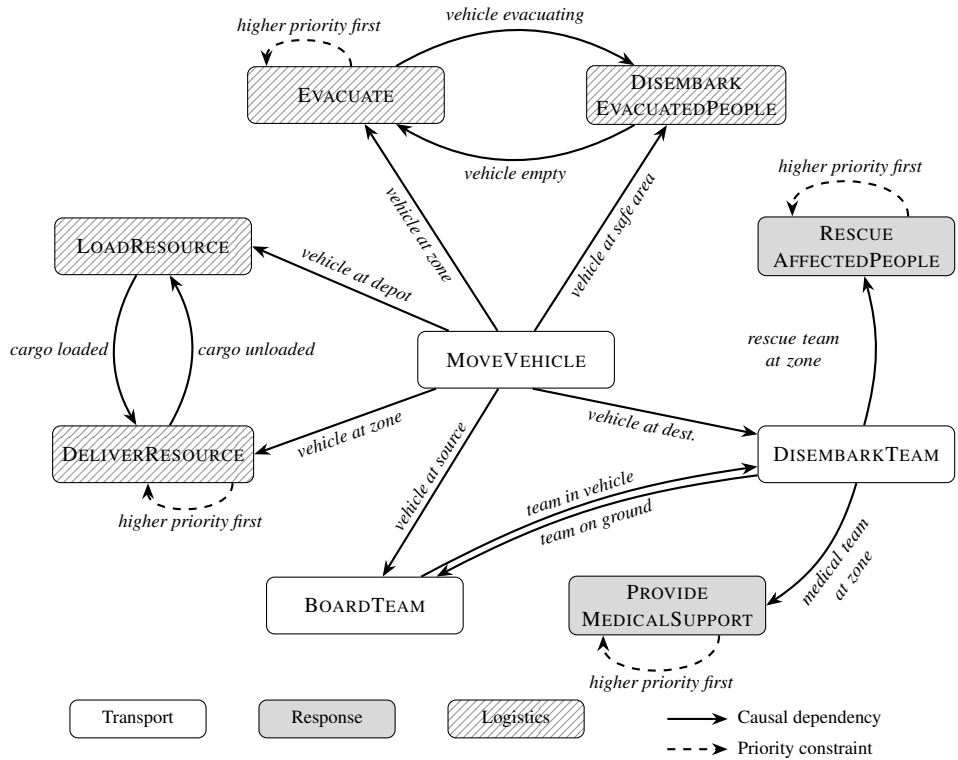
\begin{figure*}[htb]
\centering
\vspace{2em}
\begin{tikzpicture}[
  act/.style={
    draw,
    rounded corners=3pt,
    minimum width=2.6cm,
    minimum height=0.75cm,
    align=center,
    font=\footnotesize,
    text=black
  },
  tarr/.style={-{Stealth[length=6pt]}, thick},
  darr/.style={-{Stealth[length=6pt]}, thick, dashed},
  lbl/.style={font=\scriptsize\itshape, inner sep=2pt},
  transport/.style={act, fill=white},
  response/.style={act, fill=black!15},
  logistics/.style={
    act,
    fill=white,
    pattern=north east lines,
    pattern color=black!40
  }
]
  \node[transport] (board)    at (3, -.5)    {\textsc{BoardTeam}};
  \node[transport] (move)     at (5, 2.75)    {\textsc{MoveVehicle}};
  \node[transport] (disemb)   at (9.5, 1.50)   {\textsc{DisembarkTeam}};
  \node[logistics] (evacuate) at (2.75, 6.25)    {\textsc{Evacuate}};
  \node[logistics] (disembev) at (7.25, 6.25)    {\textsc{Disembark}\\ \textsc{EvacuatedPeople}};

  \node[response] (rescue)  at (9.5, 4.25) {\textsc{Rescue}\\\textsc{AffectedPeople}};
  \node[response] (medical) at (7, -0.5) {\textsc{Provide}\\\textsc{MedicalSupport}};

  \node[logistics] (load)    at (0.2, 4.25) {\textsc{LoadResource}};
  \node[logistics] (deliver) at (0.2, 1.50) {\textsc{DeliverResource}};

  \draw[tarr] (move) -- node[lbl, midway, sloped, above]{vehicle at source} (board);
  \draw[tarr] (move.300) -- node[lbl, midway, sloped, below]{vehicle at dest.} (disemb);

  \draw[tarr, bend right=15] (disemb) to node[lbl, midway, left, text width=2cm, align=center]{rescue team\\at zone} (rescue);

  \draw[tarr, bend left=20] (disemb.270) to node[lbl, midway, sloped, below, text width=2cm, align=center]{medical team\\ at zone} (medical.0);

  \draw[tarr, bend left=10] (board) to node[lbl, midway, sloped, above]{team in vehicle} (disemb);

  \draw[tarr, bend right=10] (disemb.192) to node[lbl, midway, sloped, below]{team on ground} (board.22);

  \draw[darr, bend right=90] (evacuate.80)
    to node[lbl, midway, sloped, above]{higher priority first}
    (evacuate.160);

  \draw[darr, bend right=90] (rescue.30)
    to node[lbl, midway, sloped, above]{higher priority first}
    (rescue.155);

  \draw[darr, bend left=90] (medical.330)
    to node[lbl, midway, sloped, below]{higher priority first}
    (medical.-155);

  \draw[darr, bend left=90] (deliver.340)
    to node[lbl, midway, sloped, below]{higher priority first}
    (deliver.260);

  \draw[tarr, bend left=30] (disembev) to node[lbl, midway, below]{vehicle empty} (evacuate);
    
  \draw[tarr, bend left=30] (evacuate) to node[lbl, midway, above]{vehicle evacuating} (disembev);

  \draw[tarr] (move) -- node[lbl, midway, sloped, below]{vehicle at zone} (evacuate);

  \draw[tarr] (move) -- node[lbl, midway, sloped, below]{vehicle at safe area} (disembev);

  \draw[tarr] (move) -- node[lbl, midway, sloped, above]{vehicle at depot} (load.0);

  \draw[tarr, bend right=30] (deliver) to node[lbl, midway, right]{cargo unloaded} (load);
   
  \draw[tarr, bend right=30] (load) to node[lbl, midway, left]{cargo loaded} (deliver);

  \draw[tarr] (move) -- node[lbl, midway, sloped, below]{vehicle at zone} (deliver.0);

  \begin{scope}[shift={(0,-2)}]
    \node[transport, minimum width=1.8cm, minimum height=0.5cm] at (0,0)
      {\scriptsize Transport};

    \node[response, minimum width=1.8cm, minimum height=0.5cm] at (2.5,0)
      {\scriptsize Response};

    \node[logistics, minimum width=1.8cm, minimum height=0.5cm] at (5,0)
      {\scriptsize Logistics};

    \draw[tarr] (7,0) -- (7.8,0)
      node[right, font=\scriptsize]{Causal dependency};

    \draw[darr] (7,-0.4) -- (7.8,-0.4)
      node[right, font=\scriptsize]{Priority constraint};
  \end{scope}

\end{tikzpicture}
\caption{Dependency among the actions. Transport actions are shown in plain white, response actions in light gray, and logistics actions with a diagonal hatch pattern. Solid arrows indicate causal dependencies; however, the required fact may also be available in the initial configuration. Dashed arrows indicate priority constraints.}
\label{fig:dependency2}
\end{figure*}

The actions form a structured causal network in which the effects of one action establish the preconditions of others.
This dependency structure is summarized in Figure~\ref{fig:dependency2}. 
Solid arrows denote causal dependencies, where the source action produces a fact required by the target action. Nonetheless, the target action need not depend on the source action in every execution, since the required fact may already hold in the initial configuration.
Dashed arrows represent priority constraints that prevent execution in lower-priority zones until higher-priority zones are served by the same action.
Three interacting operational cycles emerge from this structure. 
\begin{itemize}
    \item \textbf{Deployment Cycle}: moves teams from the safe hub into affected zones via vehicle boarding, movement, and disembarkation, enabling on-site response.
    \item \textbf{Evacuation Cycle}: alternates between loading civilians, transporting them to safety, and disembarking them at secure locations.
    \item \textbf{Supply Cycle}: manages the flow of relief goods from depots to affected zones through repeated loading and delivery operations.
\end{itemize}
The \textsc{MoveVehicle} action participates in all three cycles and therefore constitutes the primary shared scheduling resource in the domain. Priority constraints are enforced across response and logistics actions, ensuring strict triage ordering across zones.
\subsection{Instance Formulation}
The domain specification defines the available object types, predicates, and action schemas, but it does not determine which concrete entities participate in a particular disaster-response scenario. This information is provided by a problem instance, which specifies the initial state ($I_f$) and goal condition ($G_f$) of the planning task ($\Pi_f$) in Definition~\ref{def:flood}.

A problem instance instantiates the domain by declaring a finite set of objects, assigning values to all static predicates and lookup relations (e.g., route accessibility and travel times), and defining the initial distribution of resources, vehicles, response teams, and disaster-response demands. It also specifies the desired goal condition, thereby determining the objectives that must be achieved by a valid plan.

Table~\ref{tab:instance} specifies the planning problem corresponding to the representative flood-response scenario in Figure~\ref{fig:example}, instantiating the objects, initial state ($I_f$), and goal condition ($G_f$) of Definition~\ref{def:flood}.
\begin{table*}[!htb]
\centering
\caption{Planning task for the instance of Figure~\ref{fig:example}, instantiating the objects, initial state ($I_f$), and goal ($G_f$) of Definition~\ref{def:flood}.
}
\label{tab:instance}
\small
\setlength{\tabcolsep}{4pt}
\renewcommand{\arraystretch}{1.2}
\begin{tabular}{@{} l p{0.78\textwidth} @{}}
\toprule
\textbf{Category} & \textbf{Predicates / values} \\
\midrule
\multicolumn{2}{@{}l}{\textbf{Objects}} \\
\quad Location & \texttt{S} (SafeLocation); \texttt{A}, \texttt{B} (AffectedLocation) \\
\quad Vehicle  & \texttt{T} (TransitVehicle); \texttt{F} (FreightVehicle) \\
\quad Team     & \texttt{R} (RescueTeam); \texttt{M} (MedicalSupportTeam) \\
\quad Resource  & \texttt{food} \\
\quad PersonCount  & \texttt{p0, p20, p40, p60} \\
\quad PackageCount  & \texttt{k0, k100, k200, k300} \\
\midrule
\multicolumn{2}{@{}l}{\textbf{Initial state} ($I_f$)} \\
\quad Deployment & \texttt{vehicle\_at(T,S)}, \texttt{vehicle\_at(F,S)}, \texttt{team\_at(R,S)}, \texttt{team\_at(M,S)} \\
\quad Priority   & \texttt{prior(B,A)} \\
\quad Demands    & \texttt{needs\_rescue(A)}, \texttt{needs\_medical\_support(A)},\newline \texttt{resource\_needed(A,food,k200)}\\
\quad Status & \texttt{evac\_level(A,p60)}, \texttt{evac\_level(B,p20)} \\
\quad Supplies   & \texttt{resource\_available(S,food,k300)} \\
\quad Static     & \texttt{travel\_time(T,S,A)}{=}20, \texttt{travel\_time(T,S,B)}{=}30,\newline \texttt{travel\_time(F,S,A)}{=}25, \texttt{travel\_time(F,S,B)}{=}40,\newline
\texttt{\dots} \textcolor{gray}{// Static \texttt{team\_size}, \texttt{person\_count\_diff}, \texttt{package\_count\_diff}, \texttt{valid\_evac\_step}, and \texttt{valid\_supply\_step} values are omitted for brevity}\\
\midrule
\multicolumn{2}{@{}l}{\textbf{Goal} ($G_f$)} \\
\quad Demands & $\neg$\texttt{needs\_rescue(A)}, $\neg$\texttt{needs\_medical\_support(A)},\newline \texttt{resource\_needed(A,food,k0)} \\ 
\quad Status & \texttt{evac\_level(A,p0)}, \texttt{evac\_level(B,p0)}, $\neg$\texttt{evacuating(T)} \\
\bottomrule
\end{tabular}
\end{table*}
The goal condition ($G_f$) is achieved only when both zones have been fully evacuated (\texttt{p0}), all rescue, medical, and resource requirements in Zone~(A) have been fulfilled, and no transit vehicle remains engaged in an ongoing evacuation operation.
Together, the domain specification and a problem instance define a
temporal planning task ($\Pi_f$) that can be solved directly by an automated planner.

\subsection{Plan Validity}
\label{sec:validity}

A solution to $\Pi_f$ is a valid temporal plan
$\pi_f = \langle(t_i: a_i)\rangle_{i=1}^{n}$ in the sense of Definition~\ref{def:temporalplan}: it must be temporally consistent, satisfy each action's \emph{at-start}, \emph{over-all}, and \emph{at-end} conditions at the corresponding instants, and the state reached after executing all actions must satisfy the goal $G_f$ of Definition~\ref{def:flood}.
Beyond these requirements, the flood-response task imposes one domain-specific condition, \emph{priority ordering}.
For every pair of affected locations $\ell, \ell'$ with $\texttt{prior}(\ell, \ell') \in I_f$, no response action may start at $\ell$ while the corresponding requirement at the higher-priority location $\ell'$ remains outstanding.
Because this is enforced as a hard \emph{at start} precondition in every affected action schema (Tables~\ref{tab:response} and~\ref{tab:logistics}) rather than as an optimisation objective, every admissible plan satisfies it by construction: no planner, regardless of its search strategy, can return a valid plan that violates it, and no post-processing is required.

\section{Dynamic Replanning}
\label{sec:replanning}

Our flood response formulation in Definition~\ref{def:flood} assumes that the instance has complete and correct information about the world. In practical flood-response scenarios, critical information such as route accessibility, infrastructure status, and the number of stranded individuals may not be known a priori. Thus, plans generated prior to deployment may become partially invalid as new information emerges during execution. To address this operational reality, the proposed framework supports dynamic replanning. Whenever a discrepancy between the planner's model and the actual environment is revealed, the current planning problem is updated, and a new schedule is generated from the latest observable state. This enables the framework to adapt ongoing operations while preserving actions that have already been executed.

\begin{algorithm*}
\caption{Dynamic Replanning Procedure}
\label{alg:dynamic_replanning}
\begin{algorithmic}[1]
\Require Original Instance $\mathcal{I}$, Original Plan $\pi_0$, Surprise Time $T$, Fact Changes $\mathcal{C}$, Goal Changes $\mathcal{G}$, Planner $\mathcal{P}$
\Ensure Replanned Schedule $\pi_r$

\State $S \gets \Call{ParseInitialState}{\mathcal{I}}$
\State $\pi_0^< \gets \Call{SortByTime}{{a \in \pi_0 \mid a.\text{time} < T}}$

\Statex \Comment{\textbf{Phase 1: Reconstruct execution state}}
\For{\textbf{each} action $a \in \pi_0^<$}
\State $S \gets \Call{ApplyActionEffects}{S,a}$
\EndFor

\Statex \Comment{\textbf{Phase 2: Incorporate newly observed information}}
\State $S \gets \Call{UpdateState}{S,\mathcal{C}}$
\State $G \gets \Call{UpdateGoals}{\mathcal{I}.\text{goal},\mathcal{G}}$

\Statex \Comment{\textbf{Phase 3: Construct replanning problem}}
\State $\mathcal{I}'.\text{header} \gets \mathcal{I}.\text{header}$
\State $\mathcal{I}'.\text{objects} \gets \mathcal{I}.\text{objects}$
\State $\mathcal{I}'.\text{init} \gets \Call{ConstructInitFacts}{S}$
\State $\mathcal{I}'.\text{goal} \gets G$

\Statex \Comment{\textbf{Phase 4: Generate adapted schedule}}
\State $\pi_r \gets \Call{Plan}{\mathcal{P},\mathcal{I}'}$

\State \Return $\pi_r$
\end{algorithmic}
\end{algorithm*}

Algorithm~\ref{alg:dynamic_replanning} represents the replanning process. Given an original planning instance $\mathcal{I}$ and its corresponding plan $\pi_0$, a surprise occurring at time $T$ triggers state reconstruction by simulating all actions whose start times precede $T$. Newly observed facts and revised goals are then incorporated into the reconstructed state to produce an updated planning instance. Finally, the temporal planner is invoked on the updated instance to generate a revised schedule $\pi_r$ that reflects the newly observed conditions.

To evaluate the scalability of the proposed domain, we generated benchmarks of increasing complexity. The benchmark-generation method and complexity characteristics are described in the following section.

\section{Benchmark Generation and Replanning Setup}
\label{sec:benchmark}

This section describes the construction of both the primary benchmark used for evaluating planning performance and the replanning benchmark used to assess adaptability under dynamic environmental changes. 

\subsection{Primary Benchmark Generation}
We design a problem benchmark consisting of 50 problem instances to evaluate the scalability, runtime performance, and plan quality of the proposed domain. Each instance is encoded in PDDL 2.1 and ANML to evaluate across fundamentally different temporal planning paradigms: PDDL-based heuristic search planners and ANML-based timeline-centric planners.

The instances are organized into five tiers of increasing complexity, as summarized in Table~\ref{tab:benchmark}. Each tier consists of 10 problems. All the instances from 1 to 50 correspond to an increase in the complexity of the flood instance scenario. 
The number of grounded action spaces reaches as large as 84,579 actions. 
This reflects the combinatorial nature of the binding problem. As the number of agents and locations increases, the number of possible action groundings grows multiplicatively rather than additively.

\begin{table}[!htbp]
\centering
\caption{Characteristics of instances grouped by complexity tier; each tier includes 10 problem instances, for a total of 50.}
\label{tab:benchmark}
\small
\begin{tabular}{@{}c c c c c@{}}
\toprule
\textbf{Tier} &
\textbf{Locations} &
\textbf{Vehicles} &
\textbf{Teams} &
\textbf{Estimated Actions} \\
\toprule
1 & 2 -- 6   & 2 -- 3  & 2 -- 5   & 60 -- 554 \\
2 & 7 -- 11  & 3 -- 6  & 5 -- 7   & 632 -- 2560 \\
3 & 12 -- 15 & 6 -- 8  & 7 -- 10  & 2917 -- 7190 \\
4 & 17 -- 25 & 8 -- 12 & 10 -- 13 & 8566 -- 30632 \\
5 & 26 -- 32 & 12 -- 17 & 14 -- 17 & 32570 -- 84579 \\
\bottomrule
\end{tabular}
\end{table}

\subsection{Replanning Benchmark Generation}
To evaluate the effectiveness of the proposed replanning approach, we generated systematically perturbed instances from the PDDL instances of the primary benchmark. 
We model a single re-planning step as the regeneration of a plan from the world state observed at the moment a discrepancy is revealed.

\subsubsection{Surprise Time Selection and Perturbation Types}
Let $\pi_0$ denote the original plan, and the \emph{surprise time} $T$ is fixed at the start time of a random action near the midpoint of the plan. Placing the surprise at the midpoint ensures that a substantial portion of the plan has already been committed before the disruption occurs, while enough remains for the perturbation to demand meaningful recovery.
One of two perturbation types is injected into each instance to trigger the need for re-planning:

\begin{itemize}[leftmargin=*]
    \item \textbf{Route Blockage:}
    A route-accessibility fact is removed for all vehicles. Before committing the removal, a reachability check verifies that every \texttt{AffectedLocation} remains reachable from at least one \texttt{SafeLocation} in the residual undirected graph. Instances where blocking would disconnect the graph fall back to the evacuee increase perturbation.
    \item \textbf{Evacuee Increase:}
    The \texttt{evac\_level} of a randomly affected zone is raised to a randomly selected, strictly higher level drawn from the set of evacuation milestones. The goal condition is modified to require the zone to reach \texttt{p0}.
\end{itemize}

\subsection{Replanning Instance Preparation}
We only used 49 solved instances from the primary PDDL benchmark for the replanning experiment. Among those, odd-indexed instances are selected for the {route blockage} perturbation strategy, while even-indexed instances are selected for the {evacuee increase} perturbation strategy.
For each instance from the primary PDDL benchmark, Algorithm~\ref{alg:replan_benchmark} is applied along with the corresponding plan, random surprise time, and changes induced by the selected perturbation. 
One odd-indexed instance failed the required connectivity check and was reclassified as an evacuee increase, yielding a final distribution of 25 evacuee increase and 24 route blockage perturbations.

\begin{algorithm*}
\caption{Replanning Instance Generation}
\label{alg:replan_benchmark}
\begin{algorithmic}[1]
\Require Instance $\mathcal{I}$, Original plan $\pi_0$, surprise time function $\mathcal{T}(\cdot)$, perturbations $\mathcal{P}$
\Ensure Replanning instance $\mathcal{I}'$

\State $T \gets \mathcal{T}(\pi_0)$ \Comment{\textbf{Select surprise time near plan midpoint}}
\State $S \gets \Call{ReconstructState}{\mathcal{I}, \pi_0, T}$

\State $p \gets \Call{Sample}{\mathcal{P}}$ \Comment{\textbf{Select perturbation type}}
\If{$p = \text{Route Blockage}$}
\State $S \gets \Call{RemoveRouteFacts}{S}$
\State \textbf{if} disconnected($S$) \textbf{then} $p \gets \text{Evacuee Increase}$
\EndIf

\If{$p = \text{Evacuee Increase}$}
\State $S \gets \Call{IncreaseEvacuationDemand}{S}$
\State $G \gets \Call{UpdateGoal}{\mathcal{I}.G}$
\Else
\State $G \gets \mathcal{I}.G$
\EndIf

\State $\mathcal{I}' \gets \Call{ConstructInstance}{S, G}$
\State \Return $\mathcal{I}'$
\end{algorithmic}
\end{algorithm*}

%% file: experiment.tex
\section{Experimental Results and Analysis}
This section evaluates the results of two automated planners on both the primary benchmark and the replanning benchmark of the proposed domain. The two planners are FAPE and TFD. The evaluation addresses several questions concerning both initial plan generation and plan adaptation during execution: i) How reliably does each planner find solutions across instances of increasing complexity? ii) When both planners succeed, how do the resulting plans compare in quality, both in terms of the number of steps and the total time required to execute them? iii) How does each planner's runtime grow as the problems get larger, and at what point does growth become impractical? iv) When unexpected changes occur during execution, how much effort is required to revise existing plans, and how does the quality of the plans differ? This section addresses the findings.

\subsection*{Evaluation Metrics}
Three standard metrics are used in the primary evaluation: Coverage, Runtime, and Plan quality. 
\begin{itemize}[leftmargin=*]
    \item \textbf{Coverage} is defined as the fraction of benchmark instances for which the planner returns a valid plan within the time limit. 
    \item \textbf{Runtime} is the time from the start of the search until the planner returns a result, measured in seconds. 
    \item \textbf{Plan quality} is assessed along two dimensions: \emph{plan length} and \emph{makespan}. Plan length is the number of individual actions in the returned plan, and makespan (Definition~\ref{def:makespan}) is the total time required to execute it.
\end{itemize}
Let $\pi_0^<$ denote the subsequence of $\pi_0$ containing all actions with start time $<T$, and $\pi_r$ denote the plan produced for the corresponding re-planning instance, then--
\begin{itemize}[leftmargin=*]
  \item $M_r = \mathrm{makespan}(\pi_0^<) + \mathrm{makespan}(\pi_r)$: \emph{Total replan makespan}, combining the makespan already spent before $T$ with the plan $\pi_r$'s own makespan.
  \item $\Delta M = (M_r - M_0)/M_0$: \emph{Makespan change} relative to the original plan's makespan $M_0 = \mathrm{makespan}(\pi_0)$.
  \item $\Delta|\pi|$: \emph{Plan-length change}, comparing the length of $\pi_0$ against the combined length of the committed pre-surprise actions ($\pi_0^<$) and the plan $\pi_r$.
  \item $r_{\mathrm{rt}}$: \emph{Runtime ratio}, re-plan time divided by original planning time.
\end{itemize}

The experiments were conducted on a machine equipped with a 12th Gen Intel Core i7-12700 CPU, 16GB of 3200 MT/s RAM, and an M.2 NVMe SSD, running Ubuntu 24.04 LTS. 
Both planners were run with a maximum time limit of 3,600 seconds. An instance is counted as solved if the planner returns a valid plan within the time limit.

\subsection{Results on Primary Benchmark}

\begin{figure*}[htbp]
\centering
\begin{tikzpicture}
\begin{axis}[
    width=\textwidth,
    height=0.5\textwidth,
    xmode=log,
    y filter/.code=\pgfmathparse{2*\pgfmathresult},
    xlabel={Runtime (s)},
    ylabel={Coverage (\%)},
    legend pos=south east,
    legend style={font=\small},
    grid=both,
    grid style={dotted, gray!50},
    ymin=0, ymax=105,
    xmin=0.001, xmax=10000,
    xtick={0.001,0.01,0.1,1,10,100,1000,10000},
    ytick={0,20,40,60,80,100},
    thick,
]

\addplot[color=black!90, mark=star, mark repeat=1, mark size=3pt, smooth,thin] coordinates {
    (0.001391,1)(0.001801,2)(0.002097,3)(0.002787,4)(0.002926,5)
  (0.003083,6)(0.003481,7)(0.003588,8)(0.005329,9)(0.005334,10)
  (0.005481,11)(0.006280,12)(0.007397,13)(0.009378,14)(0.010088,15)
  (0.010349,16)(0.011345,17)(0.014296,18)(0.014648,19)(0.017284,20)
  (0.028925,21)(0.038159,22)(0.040049,23)(0.044312,24)(0.045503,25)
  (0.046879,26)(0.055632,27)(0.056243,28)(0.066646,29)(0.092568,30)
  (0.093622,31)(0.102224,32)(0.131396,33)(0.152955,34)(0.168550,35)
  (0.267610,36)(0.352677,37)(0.363903,38)(0.432769,39)(0.655239,40)
  (0.752125,41)(1.018280,42)(1.058080,43)(1.317080,44)(1.436970,45)
  (1.733660,46)(4.097520,47)(7.031110,48)(19.582700,49)
};
\addlegendentry{TFD (PDDL)}

\addplot[color=black!90, mark=o, mark repeat=1, smooth, thin] coordinates {
    (0.421,1)(0.467,2)(0.587,3)(0.653,4)(0.731,5)
  (0.743,6)(0.748,7)(0.779,8)(1.213,9)(1.219,10)
  (1.289,11)(1.394,12)(1.595,13)(1.709,14)(1.734,15)
  (1.888,16)(2.001,17)(2.116,18)(2.124,19)(3.096,20)
  (6.378,21)(6.635,22)(7.049,23)(8.519,24)(9.242,25)
  (9.780,26)(12.694,27)(18.781,28)(23.521,29)(28.363,30)
  (44.986,31)(46.204,32)(82.307,33)(87.077,34)(112.918,35)
  (271.141,36)(305.033,37)(333.685,38)(686.735,39)(834.844,40)
  (1487.472,41)(1503.870,42)(2429.159,43)
};
\addlegendentry{FAPE (ANML)}

\end{axis}
\end{tikzpicture}
\caption{Percentage of instances solved (coverage) within a given runtime. TFD achieves coverage of 98\% within 19.583 seconds, while FAPE plateaus at 86\%.}
\label{fig:coverage}
\end{figure*}
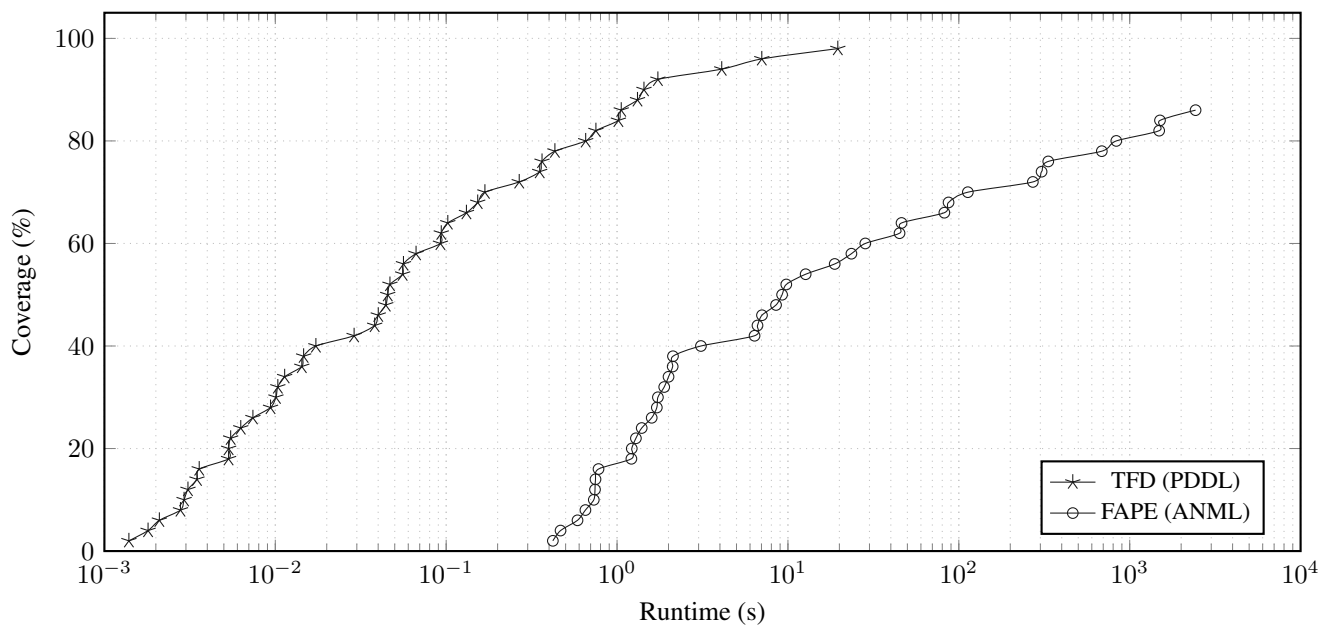

Planners' coverage and runtimes are summarized in Table~\ref{tab:results}, clustered by tiers.
\begin{table*}[htb]
\centering
\caption{Planner coverage and runtime statistics by complexity. Solved refers to instances solved within the time limit of 3,600 seconds.}
\label{tab:results}
\small
\setlength{\tabcolsep}{4pt}
\begin{tabular}{c l r r r r r}
\toprule
\multirow{2}{*}{\textbf{Tier}} &
\multirow{2}{*}{\textbf{Planner}} &
\multirow{2}{*}{\textbf{Solved}} &
\multicolumn{4}{c}{\textbf{Runtime (s)}} \\ \cmidrule(l){4-7}
& & & \textbf{Min} & \textbf{Median} & \textbf{Max} & \textbf{Mean} \\
\toprule
\multirow{2}{*}{1}
 & FAPE (ANML) & 10/10 & 0.421  & 0.737   & 1.709    & 0.806   \\
 & TFD  (PDDL) & 10/10 & 0.001  & 0.003   & 0.007    & 0.003   \\
\midrule
\multirow{2}{*}{2}
 & FAPE (ANML) & 10/10 & 1.213  & 1.811   & 3.096    & 1.845   \\
 & TFD  (PDDL) & 10/10 & 0.005  & 0.010   & 0.017    & 0.010   \\
\midrule
\multirow{2}{*}{3}
 & FAPE (ANML) &  9/10 & 6.378   & 9.242   & 28.363   & 11.938  \\
 & TFD  (PDDL) &  9/10 & 0.029   & 0.046   & 0.094    & 0.050   \\
\midrule
\multirow{2}{*}{4}
 & FAPE (ANML) &  9/10 & 23.521  & 87.077  & 333.685  & 145.208 \\
 & TFD  (PDDL) & 10/10 & 0.067   & 0.161   & 0.433    & 0.213   \\
\midrule
\multirow{2}{*}{5}
 & FAPE (ANML) &  5/10 & 686.735 & 1487.472 & 2429.159 & 1388.416 \\
 & TFD  (PDDL) & 10/10 & 0.655   & 1.377    & 19.583   & 3.868   \\
\bottomrule
\multirow{2}{*}{\textbf{All}}
 & FAPE (ANML) & \textbf{43/50} & 0.421   & 6.635   & 2429.159 & 194.951 \\
 & TFD  (PDDL) & \textbf{49/50} & 0.001   & 0.046   & 19.583   & 0.845   \\

\bottomrule
\end{tabular}
\end{table*}
Overall, TFD finds solutions faster than FAPE across tiers, and its runtime grows far more slowly as problem size increases.
Figure~\ref{fig:coverage} visualizes the cumulative coverage curve. The TFD's coverage reaches 98\% well before the 20-second mark. 
The FAPE's curve grows more slowly, plateauing at 86\% instances around 2,430 seconds.

\begin{figure*}[htbp]
\centering
\begin{tikzpicture}
\begin{loglogaxis}[
    width=\textwidth,
    height=0.5\textwidth,
    xlabel={TFD Runtime (s)},
    ylabel={FAPE Runtime (s)},
    legend pos=north west,
    legend style={font=\small},
    grid=both,
    grid style={dotted, gray!40},
    xmin=0.001, xmax=10,
    ymin=0.1,   ymax=5000,
]
\addplot[domain=0.001:5000, dashed, black, thin]
    {x};
\addlegendentry{Equal runtime}
 
\addplot[only marks, mark=o, mark size=2.5pt,
         color=black!80] coordinates {
  (0.001391,0.421)(0.001801,0.467)(0.003083,0.748)(0.002097,0.587)
  (0.002787,0.731)(0.003588,0.779)(0.003481,0.743)(0.002926,0.653)
  (0.005334,1.219)(0.007397,1.709)
};
\addlegendentry{Tier 1}

\addplot[only marks, mark=square, mark size=2.5pt,
         color=black!80] coordinates {
  (0.005481,1.213)(0.006280,1.394)(0.005329,1.289)(0.009378,1.595)
  (0.010088,1.734)(0.014296,2.001)(0.014648,1.888)(0.011345,2.124)
  (0.017284,3.096)(0.010349,2.116)
};
\addlegendentry{Tier 2}
 
\addplot[only marks, mark=triangle, mark size=3pt,
         color=black!80] coordinates {
  (0.044312,6.635)(0.038159,8.519)(0.040049,6.378)(0.028925,7.049)
  (0.056243,9.242)(0.046879,9.780)(0.093622,18.781)(0.055632,28.363)
};
\addlegendentry{Tier 3}
 
\addplot[only marks, mark=diamond, mark size=3pt,
         color=black!80] coordinates {
  (0.066646,23.521)(0.092568,44.986)(0.102224,46.204)(0.168550,82.307)
  (0.131396,87.077)(0.363903,333.685)(0.152955,112.918)
  (0.352677,305.033)(0.432769,271.141)
};
\addlegendentry{Tier 4}

\addplot[only marks, mark=pentagon, mark size=2.5pt,
         color=black!80] coordinates {
  (0.655239,1503.870)(1.058080,834.844)(0.752125,686.735)
  (1.436970,1487.472)(1.317080,2429.159)
};
\addlegendentry{Tier 5}
 
\end{loglogaxis}
\end{tikzpicture}
\caption{Log-log scatter plot of runtime for the instances solved by both planners; shapes distinguish complexity tiers. Points above the dashed diagonal indicate cases where FAPE is slower than TFD.}
\label{fig:runtime_scatter}
\end{figure*}
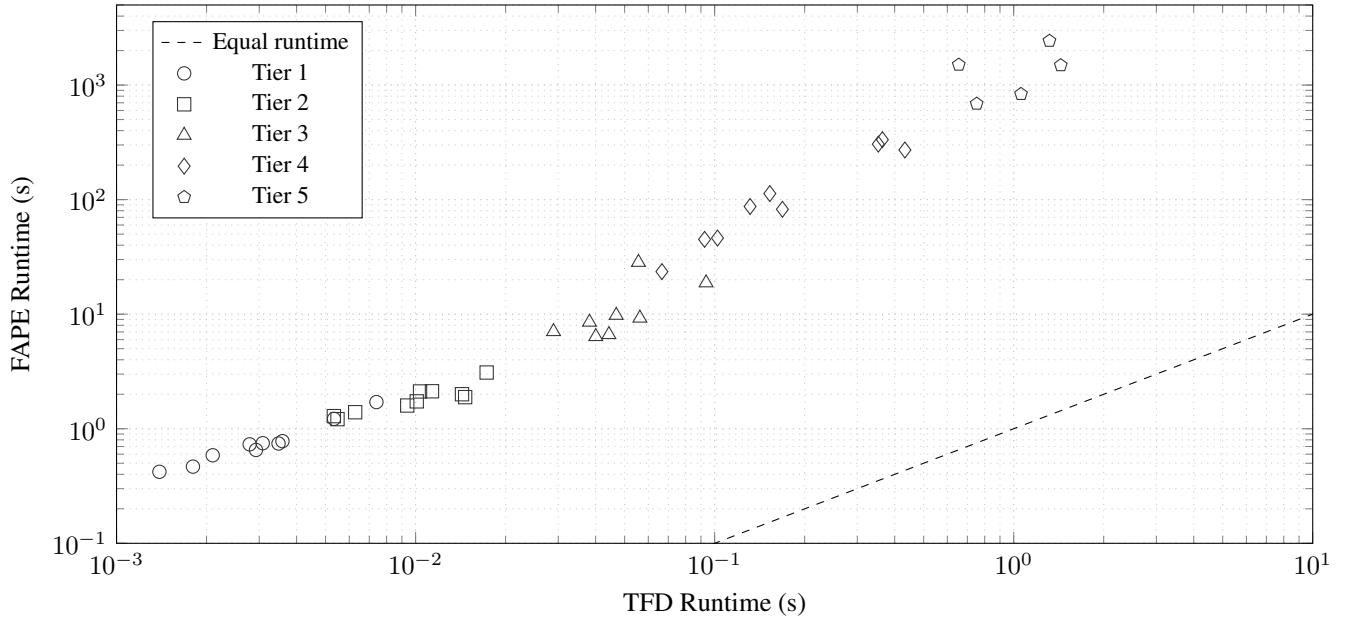

Figure~\ref{fig:runtime_scatter} shows the comparison between the planners' runtimes on the 42 instances that both solved, on a log-log scale, with points differentiated by tiers. The dashed diagonal line represents equal runtime. Every point in this plot lies above the diagonal, which confirms that TFD is consistently faster.
This difference is not surprising based on how these two planners work. TFD's heuristic-based forward search leads toward the goal faster. In contrast, the cost of FAPE's maintaining a constraint network and propagating temporal and causal constraints grows with the size of the action space.

Coverage and runtime do not directly reflect the quality of the generated plans. Table~\ref{tab:quality} presents the distribution of plan quality in terms of the plan length and makespan across the solved instances.
FAPE produces substantially shorter plans with a median of 39 steps, compared to TFD's median of 57, consistent with FAPE's planning approach. 
However, plan length is a somewhat coarse quality measure. A plan with \emph{n} steps is not worse than a shorter one if it executes steps concurrently, resulting in a smaller makespan.

\begin{table*}[!htbp]
\centering
\caption{Plan quality comparison over the instances solved by planners. Plan Length denotes the number of plan steps, and Makespan denotes the total temporal span of execution, with smaller values indicating increased parallelism.}
\label{tab:quality}
\small
\begin{tabular}{@{}l@{\hspace{10pt}}r@{\hspace{10pt}}r@{\hspace{10pt}}r@{\hspace{10pt}}r@{\hspace{10pt}}r@{\hspace{10pt}}r@{\hspace{10pt}}r@{\hspace{10pt}}r@{}}
\toprule
\multirow{2}{*}{\textbf{Planner}} & \multicolumn{4}{c}{\textbf{Plan Length}}
& \multicolumn{4}{c}{\textbf{Makespan}} \\ \cmidrule(r){2-5} \cmidrule{6-9}
 & \textbf{Min} & \textbf{Mean} & \textbf{Median} & \textbf{Max} & \textbf{Min} & \textbf{Mean} & \textbf{Median} & \textbf{Max} \\
\midrule
{\textbf{All Instances}}\\
\quad FAPE (ANML) & 14 & 49 & 39 & 140 & 170.0 & 627.4 & 533.0  & 1695.0 \\
\quad TFD  (PDDL) & 16 & 71 & 57 & 222 & 222.0 & 668.5 & 476.4  & 3223.8 \\
\midrule
{\textbf{Jointly Solved}}\\
\quad FAPE (ANML) & 14 & 49 & 39 & 140 & 170.0 & 630.6 & 536.5 & 1695.0 \\
\quad TFD  (PDDL) & 16 & 55 & 46 & 161 & 222.0 & 483.4 & 469.1  & 1519.2 \\
\bottomrule
\end{tabular}
\end{table*}

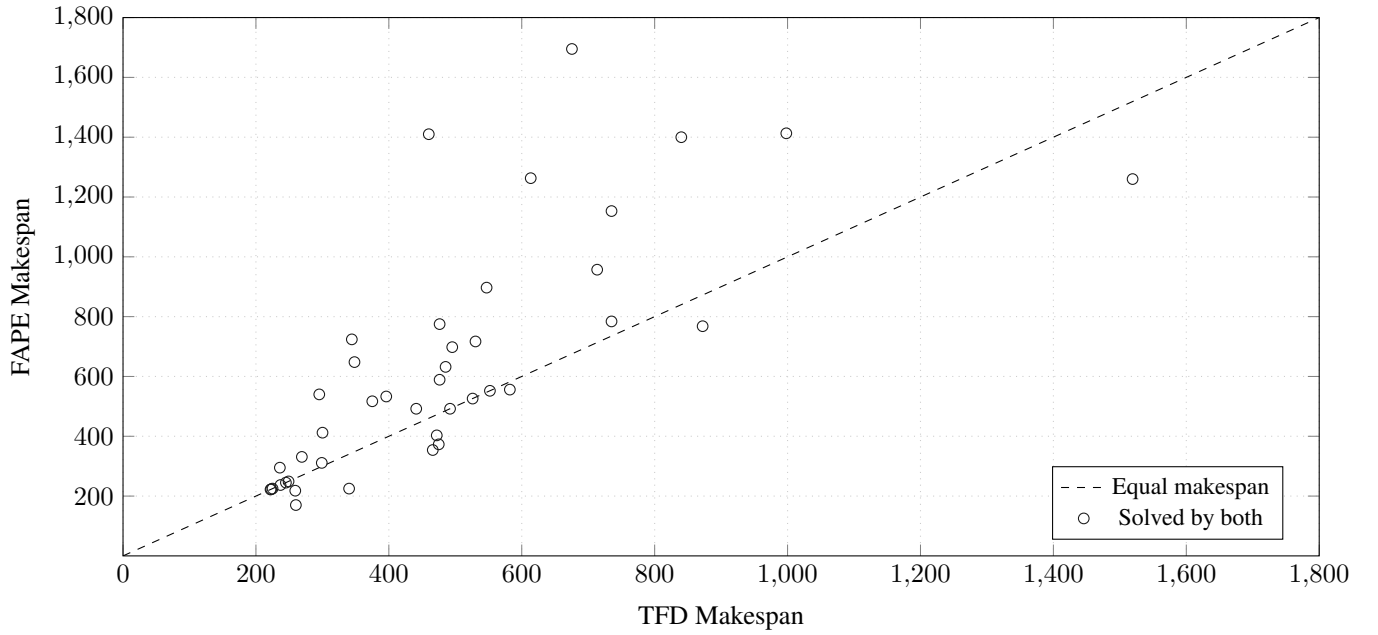
\begin{figure*}[htbp]
\centering
\begin{tikzpicture}
\begin{axis}[
    width=\textwidth,
    height=0.5\textwidth,
    xlabel={TFD Makespan},
    ylabel={FAPE Makespan},
    legend pos=south east,
    legend style={font=\small},
    grid=both,
    grid style={dotted, gray!40},
    xmin=0, xmax=1800,
    ymin=0, ymax=1800,
    ytick={200,400,600,800,1000,1200,1400,1600,1800},
]
\addplot[domain=0:1800, dashed, black, thin]{x};
\addlegendentry{Equal makespan}
\addplot[only marks, mark=o, mark size=2pt, color=black!90] coordinates {
    (526.1,526)(245.1,245)(492.1,492)(225.0,225)(299.1,311)
    (259.2,218)(222.0,222)(269.1,331)(340.2,225)(441.2,492)
    (249.1,249)(552.1,552)(237.0,237)(466.1,354)(260.1,170)
    (300.2,412)(375.1,517)(236.0,295)(475.1,373)(223.0,223)
    (530.4,717)(495.4,698)(582.1,556)(472.1,403)(485.4,632)
    (476.4,589)(613.5,1263)(396.1,533)(476.4,775)(713.5,957)
    (295.3,540)(547.1,897)(344.3,724)(460.1,1410)(348.3,648)
    (735.1,784)(872.2,768)(675.5,1695)(840.1,1400)(998.1,1413)
    (735.1,1153)(1519.2,1260)
};
\addlegendentry{Solved by both}
\end{axis}
\end{tikzpicture}
\caption{Plan makespan comparison for the 42 instances solved by both planners. Most points lie above the diagonal, indicating that FAPE generally produces plans with a higher makespan compared to TFD.}
\label{fig:makespan}
\end{figure*}

Makespan provides a more operationally meaningful measure, as it captures the total duration of execution under concurrency. Figure~\ref{fig:makespan} plots the FAPE makespan against the TFD's for all the jointly solved instances, with the diagonal line representing equal makespan. 
It shows a tendency for FAPE's makespans to be usually higher than TFD's. 
The divergence is mostly visible at the high end.
There are also several instances, points scattered below the diagonal in the lower-left region, where FAPE produces a shorter makespan than TFD.
The overall picture of plan quality is that neither planner clearly dominates the other.

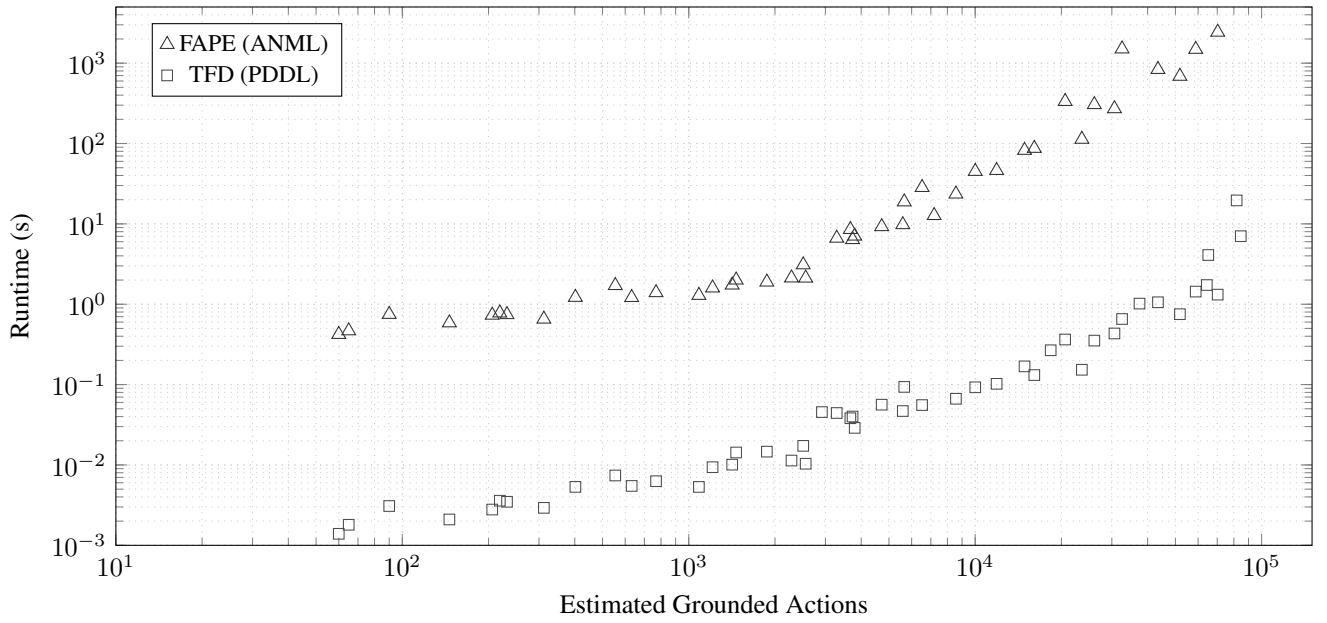
\begin{figure*}[htbp]
\centering
\begin{tikzpicture}
\begin{loglogaxis}[
    width=\textwidth,
    height=0.5\textwidth,
    xlabel={Estimated Grounded Actions},
    ylabel={Runtime (s)},
    legend pos=north west,
    legend style={font=\small},
    grid=both,
    grid style={dotted, gray!40},
    xmin=10, xmax=150000,
    ymin=0.001, ymax=5000,
]
\addplot[only marks, mark=triangle, mark size=3pt,
         color=black!80, opacity=1] coordinates {
    (60,0.421)(65,0.467)(90,0.748)(146,0.587)(206,0.731)
    (219,0.779)(232,0.743)(312,0.653)(402,1.219)(554,1.709)
    (632,1.213)(769,1.394)(1085,1.289)(1211,1.595)(1418,1.734)
    (1464,2.001)(1874,1.888)(2284,2.124)(2510,3.096)(2560,2.116)
    (3286,6.635)(3667,8.519)(3732,6.378)(3797,7.049)(4717,9.242)
    (5589,9.780)(5654,18.781)(6526,28.363)(7190,12.694)
    (8566,23.521)(10006,44.986)(11880,46.204)(14870,82.307)
    (16084,87.077)(20608,333.685)(23575,112.918)(26058,305.033)
    (30632,271.141)(32570,1503.870)(43464,834.844)(51846,686.735)
    (58868,1487.472)(70251,2429.159)
};
\addlegendentry{FAPE (ANML)}

\addplot[only marks, mark=square, mark size=2pt,
         color=black!70, opacity=1] coordinates {
  (60,0.001391)(65,0.001801)(90,0.003083)(146,0.002097)(206,0.002787)
  (219,0.003588)(232,0.003481)(312,0.002926)(402,0.005334)(554,0.007397)
  (632,0.005481)(769,0.006280)(1085,0.005329)(1211,0.009378)(1418,0.010088)
  (1464,0.014296)(1874,0.014648)(2284,0.011345)(2510,0.017284)(2560,0.010349)
  (2917,0.045503)(3286,0.044312)(3667,0.038159)(3732,0.040049)(3797,0.028925)
  (4717,0.056243)(5589,0.046879)(5654,0.093622)(6526,0.055632)
  (8566,0.066646)(10006,0.092568)(11880,0.102224)(14870,0.168550)
  (16084,0.131396)(18349,0.267610)(20608,0.363903)(23575,0.152955)
  (26058,0.352677)(30632,0.432769)(32570,0.655239)(37474,1.018280)
  (43464,1.058080)(51846,0.752125)(58868,1.436970)(64362,1.733660)
  (65130,4.097520)(70251,1.317080)(81821,19.582700)(84579,7.031110)
};
\addlegendentry{TFD (PDDL)}
\end{loglogaxis}
\end{tikzpicture}
\caption{Log-log scatter plot of the scalability of both planners with increasing problem size, measured by the size of the grounded action space.}
\label{fig:scalability}
\end{figure*}

Coverage and runtime statistics do not directly reveal how each planner behaves as the problem grows continuously. 
The scalability analysis presented in  Figure~\ref{fig:scalability} addresses this by plotting each solved instance's runtime against the size of its grounded action space, as the grounded action determines both the branching factor of the search and the size of the constraint network.

The points representing the TFD in Figure~\ref{fig:scalability} scatter linearly on the log-log axes. The overall growth rate is almost polynomial with respect to the size of the action space. The trend is consistent as the TFD planner relies on a heuristic to keep the effective branching factor low.
On the other hand, points representing the FAPE in Figure~\ref{fig:scalability} form a noisier scatter with noticeable variance.
This variance reflects the sensitivity of FAPE's constraint propagation to the specific structure of each problem, the order in which flaws are encountered and resolved, the depth of the causal chains, and the density of temporal constraints.

The key distinction is not only speed; it is the variability of FAPE's runtime within this range compared to TFD. The runtime of TFD at any given problem size is polynomial and predictable, but the runtime of FAPE is not. For operational deployment, a predictable runtime is as important as a fast runtime, because an emergency planner is more reliable if it finishes tasks in under a few seconds.

\subsection{Results on Replanning Benchmark}
The benchmark of replanning instances encoding dynamic changes mid-execution is used as input to the TFD. 
No plan repair or reuse mechanism is used; each perturbed residual problem is solved from scratch.
TFD successfully solved all 49 replanning instances.
Table~\ref{tab:replan-aggregate} summarizes the aggregate results. 
Route blockages generally produced only modest changes in the generated plans, with mean increases of 3.2\% in makespan and 3.1\% in plan length. Evacuee increases resulted in substantially larger modifications, yielding mean increases of 15.7\% in makespan and 30.1\% in plan length.

\begin{table*}[!htbp]
\centering
\caption{Aggregate replanning statistics by perturbation type. Values are means with medians in parentheses. $\Delta M$(\%) = percentage change in total makespan; $\Delta |\pi|$(\%) = percentage change in plan length; $r_\mathrm{rt}$ = re-plan runtime divided by original runtime.}
\label{tab:replan-aggregate}
\small
\begin{tabular}{@{}l c r@{\hspace{3pt}}r r@{\hspace{3pt}}r r@{\hspace{3pt}}r @{}}
\toprule
\textbf{Perturbation} &
\textbf{Number of Instances} &
\multicolumn{2}{c}{$\Delta M$(\%)} &
\multicolumn{2}{c}{$\mathbf{\Delta |\pi|}$(\%)} &
\multicolumn{2}{c}{$\mathbf{r_{\mathrm{rt}}}$} \\
\midrule
Evacuee Increase
  & 25
  & 15.7 & (6.6)
  & 30.1 & (24.3)
  & 0.78 & (0.66) \\
Route Blockage
  & 24
  & 3.2 & (0.9)
  & 3.1 & (3.1)
  & 0.80 & (0.73) \\
\midrule
\textbf{Overall}
  & \textbf{49}
  & \textbf{9.6} & \textbf{(4.2)}
  & \textbf{16.9} & \textbf{(7.7)}
  & \textbf{0.79} & \textbf{(0.71)} \\
\bottomrule
\end{tabular}
\end{table*}

\begin{figure*}[htbp]
\centering
\begin{tikzpicture}
\begin{axis}[
    width=\textwidth,
    height=0.5\textwidth,
    xmin   = -1,  xmax = 52,
    ymin   = 0.0002, ymax = 100,
    ymode  = log,
    xtick  = {1,5,10,15,20,25,30,35,40,45,50},
    xlabel = {Problem instance},
    ylabel = {Runtime (s)},
    legend style = {
        at={(0.02,0.98)}, anchor=north west,
        font=\small, draw=gray!60,
        fill=white, fill opacity=0.85
    },
    grid = both,
    grid style  = {dotted, line width=0.3pt},
    tick label style = {font=\small},
    label style = {font=\small},
]

\addplot[only marks, mark=o, mark size=2.2pt] coordinates {
    (1,0.00139)(2,0.00180)(3,0.00308)(4,0.00210)
    (5,0.00279)(6,0.00359)(7,0.00348)(8,0.00293)
    (9,0.00533)(10,0.00740)(11,0.00548)(12,0.00628)
    (13,0.00533)(14,0.00938)(15,0.01009)(16,0.01430)
    (17,0.01465)(18,0.01135)(19,0.01728)(20,0.01035)
    (21,0.04550)(22,0.04431)(23,0.03816)(24,0.04005)
    (25,0.02892)(26,0.05624)(27,0.04688)(28,0.09362)
    (29,0.05563)(31,0.06665)(32,0.09257)(33,0.10222)
    (34,0.16855)(35,0.13140)(36,0.26761)(37,0.36390)
    (38,0.15296)(39,0.35268)(40,0.43277)(41,0.65524)
    (42,1.01828)(43,1.05808)(44,0.75213)(45,1.43697)
    (46,1.73366)(47,4.09752)(48,1.31708)(49,19.58270)
    (50,7.03111)
};
\addlegendentry{Original $\pi_0$}

\addplot[only marks, mark=star, mark size=2.2pt] coordinates {
    (1,0.00139)(2,0.00120)(3,0.00911)(4,0.00220)
    (5,0.00190)(6,0.00529)(7,0.00211)(8,0.00309)
    (9,0.00250)(10,0.00710)(11,0.00470)(12,0.00714)
    (13,0.00672)(14,0.00485)(15,0.00746)(16,0.01318)
    (17,0.00861)(18,0.01575)(19,0.01382)(20,0.01808)
    (21,0.03151)(22,0.02615)(23,0.01221)(24,0.03664)
    (25,0.02039)(26,0.03360)(27,0.04117)(28,0.05807)
    (29,0.03590)(31,0.01728)(32,0.02511)(33,0.07825)
    (34,0.16977)(35,0.10668)(36,0.19579)(37,0.20446)
    (38,0.20717)(39,0.15596)(40,0.30030)(41,0.30069)
    (42,0.83878)(43,0.41272)(44,0.53759)(45,0.59317)
    (46,0.83839)(47,1.37665)(48,1.26161)(49,1.10722)
    (50,1.56997)
};
\addlegendentry{Re-plan $\pi_r$}

\end{axis}
\end{tikzpicture}
\caption{Original planning runtime versus re-planning runtime (log scale) for all instances, ordered by problem index. Overall, 38 of 49 re-plans were completed faster than the original plan.}
\label{fig:replan-runtime}
\end{figure*}
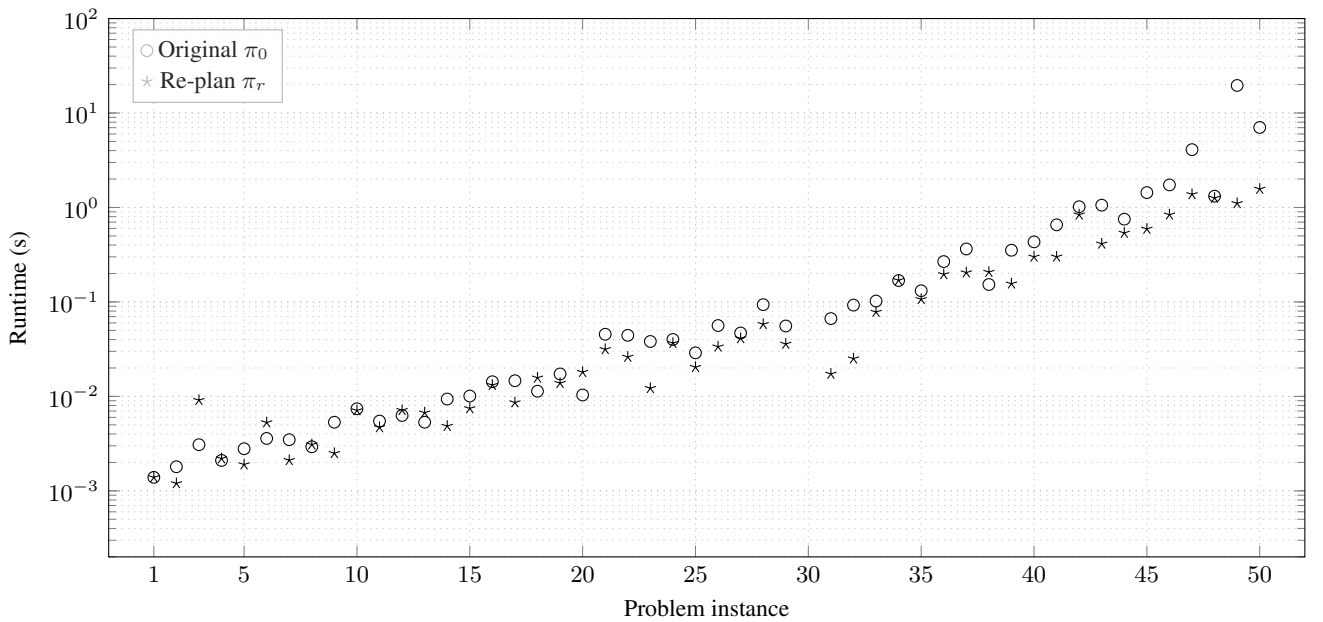

Across all instances, replanning remained computationally efficient, with a mean runtime ratio of $r_{\mathrm{rt}}=0.79$ (median $0.71$), indicating that replanning was typically faster than solving the corresponding original problem.
The comparison between original planning and replanning runtimes is presented in Figure~\ref{fig:replan-runtime} on a shared log scale.
Overall, 38 of 49 re-plans were completed in less time than the original.
The reason is straightforward: the planner inherits a partially executed state, so the subproblem it must search is smaller than the original one.

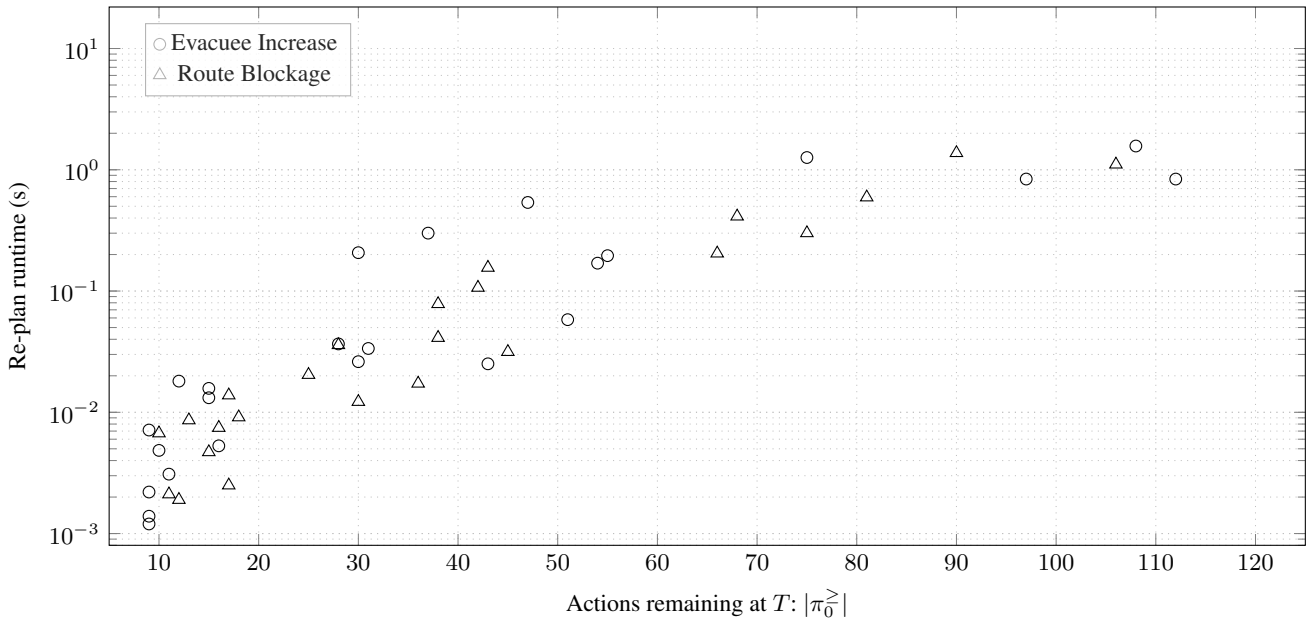
\begin{figure*}[htbp]
\centering
\begin{tikzpicture}
\begin{axis}[
    width=\textwidth,
    height=0.5\textwidth,
    xmin        = 5,   xmax = 125,
    ymin        = 0.0008, ymax = 22,
    ymode       = log,
    xlabel      = {Actions remaining at $T$: $|\pi_{0}^{\geq}|$},
    ylabel      = {Re-plan runtime (s)},
    legend style = {
        at={(0.03,0.97)}, anchor=north west,
        font=\small, draw=gray!60,
        fill=white, fill opacity=0.85,
    },
    grid = both,
    grid style  = {dotted, line width=0.3pt},
    tick label style = {font=\small},
    label style = {font=\small},
]

\addplot[only marks, mark=o, mark size=2.2pt] coordinates {
    (9,0.00139)(9,0.00120)(9,0.00220)(16,0.00529)
    (11,0.00309)(9,0.00714)(10,0.00485)(15,0.01318)
    (15,0.01575)(12,0.01808)(30,0.02615)(28,0.03664)
    (31,0.03360)(51,0.05807)(43,0.02511)(54,0.16977)
    (55,0.19579)(30,0.20717)(37,0.30030)(97,0.83878)
    (47,0.53759)(112,0.83839)(75,1.26161)(108,1.56997)
};
\addlegendentry{Evacuee Increase}

\addplot[only marks, mark=triangle, mark size=2.7pt] coordinates {
    (18,0.00911)(12,0.00190)(11,0.00211)(17,0.00250)
    (15,0.00470)(10,0.00672)(16,0.00746)(13,0.00861)
    (17,0.01382)(45,0.03151)(30,0.01221)(25,0.02039)
    (38,0.04117)(28,0.03590)(36,0.01728)(38,0.07825)
    (42,0.10668)(66,0.20446)(43,0.15596)(75,0.30069)
    (68,0.41272)(81,0.59317)(90,1.37665)(106,1.10722)
};
\addlegendentry{Route Blockage}

\end{axis}
\end{tikzpicture}
\caption{Re-planning runtime against residual plan length $|\pi_0^\geq|$ at $T$ (log-scale).  The two Re-planning costs are driven by how much of the plan remains to be reconstructed, not by what kind of perturbation triggered it.}
\label{fig:replan-runtime-vs-remaining}
\end{figure*}

Figure~\ref{fig:replan-runtime-vs-remaining} compares replanning runtime on a logarithmic scale to the number of scheduled actions remaining in $\pi_{0}^{\geq}$, which is the subsequence of $\pi_0$ containing all actions with start time $\geq$ surprise time ($T$).
A clear positive trend emerges, indicating that as the number of remaining actions grows linearly, the re-planning runtime increases exponentially.
Despite this exponential growth trajectory, the absolute runtimes remain highly efficient, proving the system's viability for real-time operational deployment.

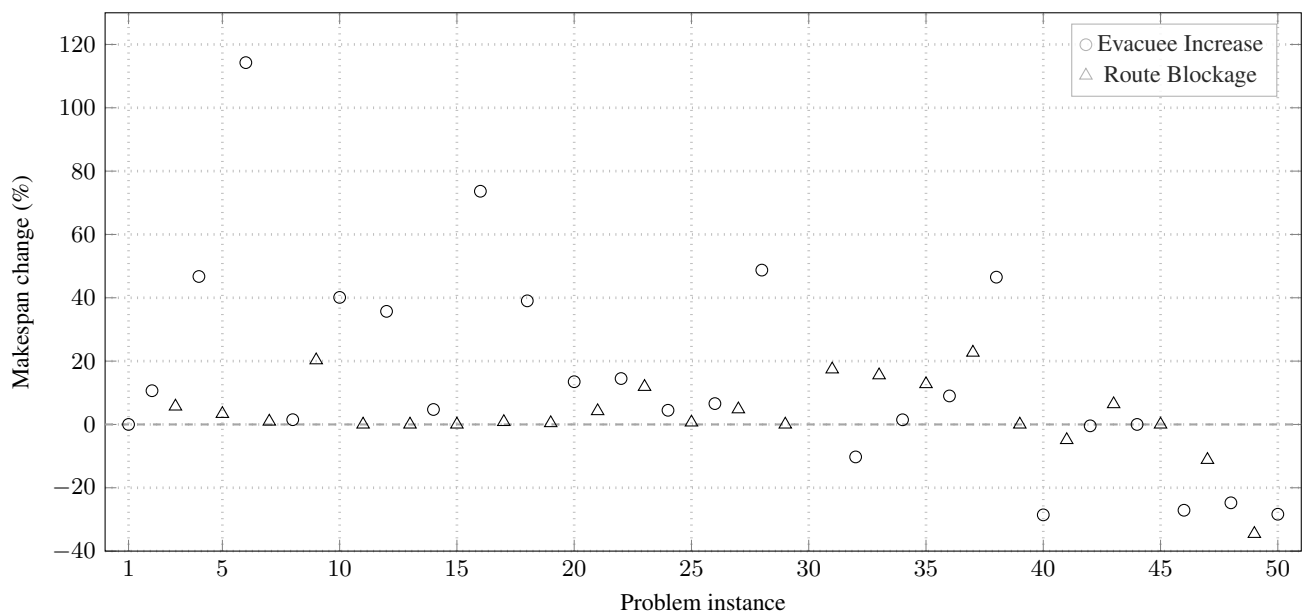
\begin{figure*}[htbp]
\centering
\begin{tikzpicture}
\begin{axis}[
    width=\textwidth,
    height=0.5\textwidth,
    xmin        = 0,  xmax = 51,
    ymin        = -40, ymax = 130,
    xtick       = {1,5,10,15,20,25,30,35,40,45,50},
    xlabel      = {Problem instance},
    ylabel      = {Makespan change (\%)},
    legend style = {
        at={(0.98,0.98)}, anchor=north east,
        font=\small, draw=gray!60,
        fill=white, fill opacity=0.85,
        legend columns=1,
    },
    grid=both,
    grid style  = {dotted, line width=1pt},
    tick label style = {font=\small},
    label style = {font=\small},
]

\draw[dashed, gray!70, line width=0.8pt]
    (axis cs:0,0) -- (axis cs:51,0);

\addplot[only marks, mark=o, mark size=2.2pt] coordinates {
    (1,-0.00)(2,10.65)(4,46.72)(6,114.26)(8,1.50)
    (10,40.12)(12,35.71)(14,4.72)(16,73.64)(18,39.03)
    (20,13.49)(22,14.49)(24,4.47)(26,6.58)(28,48.74)
    (32,-10.26)(34,1.47)(36,9.00)(38,46.51)(40,-28.56)
    (42,-0.44)(44,-0.00)(46,-27.10)(48,-24.75)(50,-28.36)
};
\addlegendentry{Evacuee Increase}

\addplot[only marks, mark=triangle, mark size=2.7pt] coordinates {
    (3,5.69)(5,3.35)(7,0.93)(9,20.29)(11,-0.00)
    (13,-0.00)(15,-0.00)(17,0.80)(19,0.44)(21,4.23)
    (23,11.89)(25,0.65)(27,4.81)(29,-0.00)(31,17.41)
    (33,15.55)(35,12.76)(37,22.70)(39,-0.00)(41,-4.90)
    (43,6.43)(45,-0.00)(47,-11.17)(49,-34.54)
};
\addlegendentry{Route Blockage}

\end{axis}
\end{tikzpicture}
\caption{Percentage change in total makespan after replanning, relative to $\pi_0$'s original makespan.  Route blockage perturbations mostly cluster near zero throughout. Evacuee increase perturbations are far more variable. The dashed horizontal line marks zero change.}
\label{fig:replan-makespan-change}
\end{figure*}

The effect of replanning on plan quality is summarized by the per-instance makespan change $\Delta M$ in Figure~\ref{fig:replan-makespan-change}.
Route blockage instances are largely unaffected: the mean increase is only $3.2\%$ (median $0.9\%$), with 19 of 24 instances below $7\%$, because rerouting typically requires only localised vehicle reassignment.

Evacuee increase perturbations are far more variable: the mean rises to $15.7\%$ with a standard deviation of $33.7\%$, reflecting the wide range of survivor-count increases imposed across instances.
Notably, several large instances yield \emph{negative} $\Delta M$: replanning from a mid-execution state allows the re-planner to discover a tighter residual schedule than the original plan's tail, resulting in a lower total makespan.

\section{Discussion}
The experimental results on the primary benchmark show that TFD is faster and more predictable than FAPE. 
The runtime gaps between the planners are large enough to matter in real-world deployment. 
Though FAPE is slower and fails on harder instances, it produces plans with fewer sequential steps, and its runtimes in small-scale instances corresponding to neighborhood-level incidents are acceptable. But for large-scale scenarios, corresponding to city-level incidents, TFD appears to be the more viable option.
FAPE's lower makespan at the smaller tiers is an advantage also.
A plan that completes in $t$ time units rather than $t'$ with $t' > t$ may correspond to dozens of additional households served before rising water levels close the last available route.

The replanning experiment under mid-execution perturbations extends the findings. Even under dynamic conditions, TFD remains highly robust and successfully solves all replanning instances.
The results further confirm that replanning is generally faster than initial planning. In some cases, replanning from a partially executed state produced plans with shorter makespans than the corresponding tails of the original plans. One possible reason is that the planner identifies alternative coordination patterns that were not selected during the initial planning phase.

For operational deployment, the most robust approach is to consider the TFD and FAPE in combination, depending on requirements.
For initial planning, the experimental results provide strong support for selecting TFD as the default planner because of its computational efficiency and consistent performance. 
FAPE is a viable choice when the scenario is small-scale, plan quality is the priority, and there is time to do better. 
Mid-execution replanning is a different situation entirely. The operation is already moving, and what matters is getting a usable plan back quickly. Under such conditions, TFD appears to be the more suitable choice.

%% file: conclusion.tex
\section{Conclusion}
Can automated planning be applied to produce temporally coherent response plans for floods? 
This was the central question addressed in this work. 
To answer it, a flood-response framework was developed that captures the full operational cycle of coordinated disaster response, including rescue, medical support, evacuation, and supply delivery. 
The proposed framework supports key operational considerations such as prioritization of affected zones, route accessibility, action concurrency, and other real-world constraints. The framework also enables mid-execution replanning under dynamic environmental changes, enhancing its applicability.
To validate the framework, experiments on initial response planning and mid-execution replanning were conducted using a benchmark set of instances of the proposed framework’s temporal planning domain.
Results indicate that automated planning can generate temporal flood-response plans while supporting dynamic adaptation during mid-execution surprises.
However, the findings are based on experiments on the benchmark instances, and additional validation using real-world flood response data and operational workflows remains necessary to assess practical deployment feasibility.

Future work could extend the proposed framework to support multi-objective optimization, such as makespan and resource efficiency, together with post-processing mechanisms that enable decision-makers to explore trade-offs among competing objectives. In addition, developing an interface that translates planner output into human-readable schedules or dispatch orders would move the proposed framework from a research prototype toward something usable in an emergency operations center.